%% file: main.tex
\documentclass{article}

\newcommand{\hideh}[1]{}

\usepackage[preprint]{neurips_2021}

\usepackage[utf8]{inputenc} 
\usepackage[T1]{fontenc}    
\usepackage{hyperref}       
\usepackage{url}            
\usepackage{booktabs}       
\usepackage{amsfonts}       
\usepackage{nicefrac}       
\usepackage{microtype}      
\usepackage{xcolor}         

\usepackage{amsthm}
\usepackage{amsmath}
\usepackage{amsfonts}
\usepackage{multirow}
\usepackage{microtype}
\usepackage{graphicx}

\usepackage{caption} 
\usepackage{subcaption} 
\usepackage{wrapfig} 

\usepackage{booktabs} 

\newtheorem{theorem}{Theorem}[section]

\newtheorem{lemma}[theorem]{Lemma}
\newtheorem{proposition}[theorem]{Proposition}

\newenvironment{hproof}{%
  \proof}{\endproof}

\title{Algorithmic Recourse in the Wild: \\
Understanding the Impact of Data and Model Shifts}

\author{%
  Kaivalya Rawal \\
  Harvard University \\
  \texttt{kaivalyarawal45@gmail.com} \\
  \And
  Ece Kamar \\
  Microsoft Research \\
  \texttt{eckamar@microsoft.com} \\
  \And
  Himabindu Lakkaraju \\
  Harvard University \\
  \texttt{hlakkaraju@hbs.edu} \\
  
  \hideh{
  \texttt{hippo@cs.cranberry-lemon.edu} \\
  David S.~Hippocampus\thanks{Use footnote for providing further information
    about author (webpage, alternative address)---\emph{not} for acknowledging
    funding agencies.} \\
  Department of Computer Science\\
  Cranberry-Lemon University\\
  Pittsburgh, PA 15213 \\
  \texttt{hippo@cs.cranberry-lemon.edu} \\}
}

\begin{document}

\maketitle

\begin{abstract}
  \hideh{The abstract paragraph should be indented \nicefrac{1}{2}~inch (3~picas) on
  both the left- and right-hand margins. Use 10~point type, with a vertical
  spacing (leading) of 11~points.  The word \textbf{Abstract} must be centered,
  bold, and in point size 12. Two line spaces precede the abstract. The abstract
  must be limited to one paragraph.}

As predictive models are increasingly being deployed to make a variety of consequential decisions, there is a growing emphasis on designing algorithms that can provide recourse to affected individuals. Existing recourse algorithms function under the assumption that the underlying predictive model does not change. However, models are regularly updated in practice for several reasons including data distribution shifts.  In this work, we make the first attempt at understanding how model updates resulting from data distribution shifts impact the algorithmic recourses generated by state-of-the-art algorithms. We carry out a rigorous theoretical and empirical analysis to address the above question. Our theoretical results establish a lower bound on the probability of recourse invalidation due to model shifts, and show the existence of a tradeoff between this invalidation probability and typical notions of ``cost" minimized by modern recourse generation algorithms. We experiment with multiple synthetic and real world datasets, capturing different kinds of distribution shifts including temporal shifts, geospatial shifts, and shifts due to data correction. These experiments demonstrate that model updation due to all the aforementioned distribution shifts can potentially invalidate recourses generated by state-of-the-art algorithms. Our findings thus not only expose previously unknown flaws in the current recourse generation paradigm, but also pave the way for fundamentally rethinking the design and development of recourse generation algorithms.
\end{abstract}

\vspace{-0.1in}
\section{Introduction}
\vspace{-0.1in} 
\input{sections/010introduction}

\vspace{-0.1in}
\section{Related Work}
\vspace{-0.1in} 
\input{sections/020relwork}

\vspace{-0.1in}
\section{Theoretical Analysis}
\vspace{-0.1in} 
\input{sections/030theory}

\vspace{-0.1in}
\section{Experimental Analysis}
\vspace{-0.1in} 
\input{sections/040experiments}

\vspace{-0.10in}
\section{Conclusion}
\vspace{-0.15in} 
\input{sections/050conclusion}

\newpage

\bibliographystyle{plainnat}
\bibliography{references}

\newpage
\newpage

\hideh{
\section*{Checklist}
}
\hideh{
The checklist follows the references.  Please
read the checklist guidelines carefully for information on how to answer these
questions.  For each question, change the default \answerTODO{} to \answerYes{},
\answerNo{}, or \answerNA{}.  You are strongly encouraged to include a {\bf
justification to your answer}, either by referencing the appropriate section of
your paper or providing a brief inline description.  For example:
\begin{itemize}
  \item Did you include the license to the code and datasets? \answerYes{See Section~\ref{gen_inst}.}
  \item Did you include the license to the code and datasets? \answerNo{The code and the data are proprietary.}
  \item Did you include the license to the code and datasets? \answerNA{}
\end{itemize}
Please do not modify the questions and only use the provided macros for your
answers.  Note that the Checklist section does not count towards the page
limit.  In your paper, please delete this instructions block and only keep the
Checklist section heading above along with the questions/answers below.
} 
\hideh{

\begin{enumerate}

\item For all authors...
\begin{enumerate}
  \item Do the main claims made in the abstract and introduction accurately reflect the paper's contributions and scope?
    \answerYes{}
  \item Did you describe the limitations of your work?
    \answerNA{We do not propose any new methods, but instead analyse existing state-of-the-art techniques. We describe potential future work in our conclusion section.}
  \item Did you discuss any potential negative societal impacts of your work?
    \answerNA{The paper itself points out negative social impacts of current state-of-the-art algorithmic recourses.}
  \item Have you read the ethics review guidelines and ensured that your paper conforms to them?
    \answerYes{}
\end{enumerate}

\item If you are including theoretical results...
\begin{enumerate}
  \item Did you state the full set of assumptions of all theoretical results?
    \answerYes{All assumptions made in theorems are listed as separate propositions/lemmas/ or stated in the proof sketches.}
	\item Did you include complete proofs of all theoretical results?
    \answerYes{The paper has sketches, but full proofs follow in the appendix.}
\end{enumerate}

\item If you ran experiments...
\begin{enumerate}
  \item Did you include the code, data, and instructions needed to reproduce the main experimental results (either in the supplemental material or as a URL)?
    \answerYes{Except for the bail dataset, which is proprietary for data privacy reasons}
  \item Did you specify all the training details (e.g., data splits, hyperparameters, how they were chosen)?
    \answerYes{The hyperparameters are fixed, and the code can be run end-to-end to replicate our results}
	\item Did you report error bars (e.g., with respect to the random seed after running experiments multiple times)?
    \answerNA{The accuracy results were averaged over 10 runs, as mentioned in the experiments section}
	\item Did you include the total amount of compute and the type of resources used (e.g., type of GPUs, internal cluster, or cloud provider)?
    \answerNo{All experiments were run on a single laptop with 16 GB ram, and an 8th gen i-7 processor. The scripts provided in the supplements should run end to end in a few hours in such a setup. This information will be included in the open source upload of this code to GitHub later.}
\end{enumerate}

\item If you are using existing assets (e.g., code, data, models) or curating/releasing new assets...
\begin{enumerate}
  \item If your work uses existing assets, did you cite the creators?
    \answerYes{\cite{lakkaraju2016interpretable, geospatial, UCI, ngcredit}}
  \item Did you mention the license of the assets?
    \answerYes{The Bail dataset is proprietary, but the others are in the public domain (Creative commons and UCI)}
  \item Did you include any new assets either in the supplemental material or as a URL?
    \answerNo{}
  \item Did you discuss whether and how consent was obtained from people whose data you're using/curating?
    \answerNA{The data was obtained without any personal information}
  \item Did you discuss whether the data you are using/curating contains personally identifiable information or offensive content?
    \answerNA{There is no such information}
\end{enumerate}

\item If you used crowdsourcing or conducted research with human subjects...
\begin{enumerate}
  \item Did you include the full text of instructions given to participants and screenshots, if applicable?
    \answerNA{}
  \item Did you describe any potential participant risks, with links to Institutional Review Board (IRB) approvals, if applicable?
    \answerNA{}
  \item Did you include the estimated hourly wage paid to participants and the total amount spent on participant compensation?
    \answerNA{}
\end{enumerate}

\end{enumerate}
}
\appendix
\input{sections/060appendix.tex}


\hideh{
\appendix

\section{Appendix}
\input{sections/050appendix.tex}
Optionally include extra information (complete proofs, additional experiments and plots) in the appendix.
This section will often be part of the supplemental material.
}
\end{document}

%% file: sections/010introduction.tex
Over the past decade, machine learning (ML) models are being increasingly deployed in diverse applications across a variety of domains ranging from finance and recruitment to criminal justice and healthcare. Consequently, there is growing emphasis on designing tools and techniques which can explain the decisions of these models to the affected individuals and provide a means for \emph{recourse} (\cite{voigt2017eu}). For example, when an individual is denied loan by a credit scoring model, he/she should be informed about the reasons for this decision and what can be done to reverse it. Several approaches in recent literature have tackled the problem of providing recourses to affected individuals by generating \emph{local} (instance level) counterfactual explanations. For  instance, ~\cite{wachter2018a} proposed a model-agnostic, gradient based approach to find a closest modification (counterfactual) to any data point which can result in the desired prediction. ~\cite{Ustun_2019} proposed an efficient integer programming approach to obtain \emph{actionable} recourses for linear classifiers. 

While prior research on algorithmic recourses has mostly focused on providing counterfactual explanations (recourses) for affected individuals, it has left a critical problem unaddressed. It is not only important to provide recourses to affected individuals but also to ensure that they are robust and reliable. It is absolutely critical to ensure that once a recourse is issued, the corresponding decision making entity is able to honor that recourse and approve any reapplication that fully implements all the recommendations outlined in the prescribed recourse (\cite{wachter2018a}). If the decision making entity cannot keep its promise of honoring the prescribed recourses (i.e., recourses become invalid), not only would the lives of the affected individuals be adversely impacted, but the decision making entity would also lose its credibility.

One of the key reasons for recourses to become invalid in real world settings is the fact that several decision making entities (e.g., banks and financial institutions) periodically retrain and update their models and/or use online learning frameworks to continually adapt to new patterns in the data. Furthermore, the data used to train these models is often subject to different kinds of distribution shifts (e.g, temporal shifts) (\cite{rabanser2019failing}). Despite the aforementioned critical challenges, there is little to no research that systematically evaluates the reliability of recourses and assesses if the recourses generated by state-of-the-art algorithms are robust to distribution shifts. While there has been some recent research that sheds light on the spuriousness of the recourses generated by state-of-the-art counterfactual explanation techniques and advocates for approaches grounded in causality (\cite{Barocas_2020, karimi2020algorithmic1, karimi2020algorithmic2, venkat}), these works do not explicitly consider the challenges posed by distribution shifts. 

In this work, we study if the recourses generated by state-of-the-art counterfactual explanation techniques are robust to model updates caused by data distribution shifts. To the best of our knowledge, this paper makes the first attempt at studying this problem. We theoretically analyze under what conditions recourses get invalidated - we prove that there is a tradeoff between recourse cost and the potential chance of invalidation due to model updates, and find a non-zero lower bound on the probability of invalidation of recourses due to distribution shifts. These key findings motivate our experiments identifying recourse invalidation in real world datasets.

We consider various classes of distribution shifts in our experimental analysis, namely: temporal shifts, geospatial shifts, and shifts due to data corrections. We deliberately select datasets from disparate domains such as criminal justice, education, and credit scoring, in order to stress the broad applicability and serious practical impacts of recourse invalidation. Our experimental results demostrate that model updates cause by all the aforementioned distribution shifts could potentially invalidate recourses generated by state-of-the-art algorithms, including causal recourse generation algorithms. We thus expose previously unknown problems with recourse generation that are broadly applicable to all currently known algorithms for generating recourses. This further emphasizes the need for rethinking the design and development of recourse generation algorithms and counterfactual explanation techniques as they stand today.

%% file: sections/020relwork.tex
\label{relwork}
A variety of post hoc techniques have been proposed to explain complex models (\cite{doshi2017towards,ribeiro2018anchors,koh2017understanding}). 
These may be model-agnostic, local explanation approaches (\cite{ribeiro2016should, lundberg2017unified}) or methods to capture feature importances (\cite{simonyan2013deep, sundararajan2017axiomatic, selvaraju2017grad,smilkov2017smoothgrad}). A different class of post-hoc local explainability techniques proposed in the literature is counterfactual explanations, which can be used to provide algorithmic recourse \footnote{Note that the literature often uses related terms such as counterfactual explanation, actionable recourse, or contrastive explanations. In this paper, we adopt a broad definition and use these terms interchangeably. For our purposes, algorithmic recourse or a counterfactual explanation consists of finding a positively classified data point, given a point that was classified negatively by a black box binary classifier.}. Separately, there has also been research theoretically and empirically analysing distribution shifts (\cite{rabanser2019failing, subbaswamy2020evaluating}) and adversarial perturbation attacks, and their consequent impacts on machine learning classifier accuracies and uncertainty (\cite{ovadia2019trust}), and interpretability and fairness (\cite{thiagarajan2020accurate}). While there has been a lot of work on distribution shifts and on recourses individually, these have remained largely separate pursuits and we wish to fill this gap in the literature by studying the behaviour of algorithmic recourses under distribution shifts.

Recourse generation algorithms have been developed for tree based ensembles (\cite{Tolomei_2017, lucic2019actionable}), using feature-importance measures (\cite{rathi2019}), perturbations in latent spaces defined via autoencoders (\cite{Pawelczyk_2020, joshi2019realistic}), SAT solvers (\cite{MACE}), genetic algorithms (\cite{Sharma_2020, schleich2021geco}), determinantal point processes (\cite{Mothilal_2020}), and the growing spheres method (\cite{laugel2017inverse}), among others. Two methods are particularly noteworthy for their generic formulations, enabling their application in diverse contexts. \cite{wachter2018a} first proposed counterfactual explanations, using gradients to directly optimize the L1 or L2 normed distance between a data point $\mathcal{S}$ and corresponding counterfactual recourse $\mathcal{S'}$. \cite{Ustun_2019} used integer programming tools to minimize the log absolute difference between a data point $\mathcal{S}$ and the recourse $\mathcal{S'}$, but only considered those feature values for $\mathcal{S'}$ that already exist in the data. This is one way of ensuring that the recourses generated are ``actionable".

In this paper we broadly adopt the classification taxonomy proposed by \cite{pred-multiplicity}; which classifies recourse generation techniques into two types: those that promote greater sparsity (\textbf{sparse counterfactuals})  (\cite{wachter2018a, laugel2017inverse, looveren2019interpretable, FACE, Tolomei_2017}), and those that promote greater support from the given data distribution (\textbf{data support counterfactuals}) (\cite{Ustun_2019, looveren2019interpretable, Pawelczyk_2020, joshi2019realistic, FACE}). We can further qualify additional recourse generation techniques that use a causal notion of data support probabilities to form an important subcategory termed (\textbf{Causal Counterfactuals}) (\cite{karimi2020algorithmic1, karimi2020algorithmic2}). \cite{pred-multiplicity} also goes on to show that there is a known \emph{upper} bound on recourse cost under predictive multiplicity. However, none of the works referenced above explicitly focus on analyzing if the recourses generated by state-of-the-art algorithms are robust to model updates caused by distribution shifts.

\hideh{
The first method proposed to generate counterfactual explanations was introduced by \cite{wachter2018a}. This method guides the search for recourse using the model gradients. Adopting the notation used in this work, we can consider a given model $f_w$ trained by minimising (for given loss function $l$ and regularisation function $\rho$) the objective $l(f_w(x_i), y_i) + \rho(w)$ to find a set of model weights $w$. Now we similarly minimise (for hyperparameter $\lambda$ and arbitrary distance metric $d$) the related objective $\lambda(f_w(x')-y')^2 + d(x_i, x')$ w.r.t $x'$ to find the counterfactual $x'$ to the sample $x$. This search technique relies on the same gradient based optimisation used to train the classifier $f_w(x)$, and ADAM is the recommended optimiser to do this. Further, using an L1 regularisation on the distance metric to induce sparsity is found to provide greater interpretability. \cite{laugel2017inverse} introduced a very similar technique too. Although they do not use the term counterfactual, their \textit{growing spheres} technique has the effect of directly generating candidate modifications in the vicinity of the data point $\mathcal{S}$. They then select the closest (sparse) data point to $\mathcal{S}$ for which the model predictions are changed as desired.

Often, it is not possible to use any generated counterfactual explanation as recourse, because it contains recommended changes that are not feasible. To ensure that recourse is actionable, we wish to restrict ourselves to those candidate modifications which are deemed realistic and possible to obtain in the real world. Many techniques use the training data $\mathcal{D}$ to gauge this feasibility, by assuming that candidate modifications $\mathcal{S'}$ are viable only if they already occur in the training data. For the limited case of linear models, \cite{Ustun_2019} propose "actionable recourse" - a technique to generate recourses using integer programming tools. This restriction on the model class is ensured by the use of monotonic continuous cost functions defined independently on each predictor input to the model $\mathcal{B}$.

\cite{looveren2019interpretable} also try to build in such feasibilty into their counterfactuals, and ensure that none of them are out of distribution. To do this they guide their search towards the single prototype instance of the desired class. They also demonstrate experimentally that there is a trade-off between inducing sparsity and feasibility in generated counterfactuals. \cite{MACE} develop MACE, which meticulously converts each trained model into an exactly equivalent boolean SAT problem using first-order-logic. Since this is theoretically feasible for any arbitrary model, their approach is model agnostic. They then use standard SAT solvers to repeatedly generate candidate modifications and test them against the classifier until a configurations with a desired output is found. \cite{FACE} develop FACE, where modifications are thought of as steps taken in feature space towards regions with favourable classifier outputs. They guide their search by selecting steps along paths with high kernel density - thus leading to counterfactuals that are more feasible due to their proximity to real world data points. This approach, like the actionable recourse technique, explicitly uses the training-data $\mathcal{D}$ when generating recourses.

Other approaches have utilized explanations output by methods such as SHAP~\cite{rathi2019},  tree based ensembles~\cite{Tolomei_2017,lucic2019actionable}, or perturbations in latent space found by autoencoders~\cite{Pawelczyk_2020,joshi2019realistic}, to generate recourses. More recently, there have also been proposals to use causal data generation models in order to ensure that recourses that are more likely to be valid in real world use cases \cite{Barocas_2020, karimi2020algorithmic1, karimi2020algorithmic2}. These works use synthetically generated data in their analyses and leverage simple examples to show that causal counterfactual explanations are more robust than those found through statistical means.

Distribution shifts are naturally occurring phenomena which have a large impact on different areas of machine learning practice and research. \cite{ovadia2019trust} study the effects of distribution shifts on predictive uncertainty, focusing on accuracy and calibration. Similarly, \cite{subbaswamy2020evaluating} show how a the robustness of a model to distribution shifts can be measured effectively. \cite{rabanser2019failing} further go on to explore empirical methods that can identify real world distribtuion shifts. Finally, \cite{thiagarajan2020accurate, DBLP:journals/corr/abs-1911-00677} are able to analyze the impacts of distribution shifts on feature importance explainability and the identification and mitigation of fairness violations.

\subsection{Types of Recourses}

Before proceeding further, we adopt the classification taxonomy proposed by \cite{pred-multiplicity}; which broadly classified recourse generation techniques into two types: those that promote greater sparsity (\textbf{sparse counterfactuals})  \cite{wachter2018a, laugel2017inverse, looveren2019interpretable, FACE, Tolomei_2017}), and those that promote greater support from the given data distribution (\textbf{data support counterfactuals}) \cite{Ustun_2019, Pawelczyk_2020, laugel2018defining, joshi2019realistic, MACE, karimi2020algorithmic1, karimi2020algorithmic2}). For a given set of counterfactual explanations $\mathcal{S'}: \mathcal{S'} \in \mathcal{B}(\mathcal{S'})=+1$ and arbitrary cost function $d(\mathcal{S}, \mathcal{S'})$ (usually set to $d(\mathcal{S}, \mathcal{S'}) = ||\mathcal{S}-\mathcal{S'}||_p$, $p \in \{1, 2\}$) these correspond to counterfactual explanations defined as $arg min_{\mathcal{S'}} {d(\mathcal{S'}, \mathcal{S})}$ and $arg min_{\mathcal{S'}: P_{data}(\mathcal{S'})>0} {d(\mathcal{S'}, \mathcal{S})}$ respectively. It is further shown here that there necessarily exists a tradeoff between these two notions of cost, and that there is an upper bound on the cost under predictive multiplicity. However, none of these works explicitly focus on analyzing if the recourses generated by state-of-the-art algorithms are robust to distribution shifts.

}

%% file: sections/030theory.tex
In this section we first define our notation, and provide proof sketches for our theoretical results. Detailed proofs can be found in the appendix.

\vspace{-0.05in}
\subsection{Preliminaries}
\vspace{-0.1in}

We consider a black box binary classifier $\mathcal{M}$ trained on a dataset of real world users $\mathcal{D}$ and for each user $\mathcal{S}$ in $\mathcal{D}$, let us consider the corresponding recourse found by a recourse generation algorithm $Alg$ to be $\mathcal{S'}$.

Further, let the set of adversely (negatively) classified users be $\mathcal{D}^{M-}$, positively classified users be $\mathcal{D}^{M+}$, and the set of all recourses be $\mathcal{CF}_1$. Our definitions can be thus written as follows:

\vspace{-0.17in}

\begin{align}
 \mathcal{D}^{M+} &= \{\mathcal{S} \in \mathcal{D} : \mathcal{M}(\mathcal{S}) = +1\} \\
 \mathcal{D}^{M-} &= \{\mathcal{S} \in \mathcal{D} : \mathcal{M}(\mathcal{S}) = -1\} \\
 \mathcal{CF}_1 &= \{ \mathcal{S} \in \mathcal{D}^{M-}, \mathcal{S'} : Alg(\mathcal{S}) = \mathcal{S'} \} 
\end{align}
\vspace{-0.15in}

\hideh{
From our construction the following three statements always hold:
\begin{align}
\mathcal{M}(\mathcal{S}) &= -1 \;\; \forall  \;\; \mathcal{S} \in \mathcal{D}^{M-} \\
\mathcal{M}(\mathcal{S}) &= +1 \;\; \forall  \;\; \mathcal{S} \in \mathcal{D}^{M+} \\
\mathcal{M}(\mathcal{S'}) &= +1 \;\; \forall  \;\; \mathcal{S} \in \mathcal{CF}_1 
\end{align}
} 

By definition, $\mathcal{M}(\mathcal{S'}) = +1 \;\; \forall  \;\; \mathcal{S} \in \mathcal{CF}_1$. All recourse generation methods essentially perform a search starting from a given data point $\mathcal{S}$, going towards $\mathcal{S'}$, by repeatedly polling the input-space of the black box model $\mathcal{M}$ along the path $\mathcal{S} \rightarrow \cdots \mathcal{S"} \cdots \rightarrow \mathcal{S'}$. There are many ways in which to guide this path when searching for counterfactual explanations. Sparse recourse generation techniques (\cite{Pawelczyk_2020}) operate on a given data point $\mathcal{S}$, and an arbitrary cost function $d(\mathcal{S}, \mathcal{S'})$ (usually set to $d(\mathcal{S}, \mathcal{S'}) = ||\mathcal{S}-\mathcal{S'}||_p$, $p \in \{1, 2\}$). For example, \cite{wachter2018a} define optimal recourses according to eqn \ref{eqn:wachter}. Data support recourses restrict the counterfactual search to those $\mathcal{S'}$ that are found in the data distribution (with high probability), as in eqn \ref{eqn:berk}. \cite{Ustun_2019} use a heuristic to try and optimise this using linear programming tools. \label{sec:prelims}

\vspace{-0.17in}

\begin{align}
    \mathcal{S'} &= \arg \min_{\mathcal{S'}} \, {d(\mathcal{S}, \mathcal{S'})} \label{eqn:wachter} \\
    \mathcal{S'} &= \arg \min_{\mathcal{S'}: P_{data}(\mathcal{S'})>0} \, {d(\mathcal{S}, \mathcal{S'})} \label{eqn:berk}
\end{align}

\vspace{-0.15in}

\vspace{-0.05in}
\subsection{Our Setup and Assumptions}
\vspace{-0.1in}
We consider a standardized setup to compare distribution shifts in different scenarios. This is designed to be similar to real-world machine learning deployments, where models deployed in production are often updated by retraining periodically on new data. We consider the following setup to analyze and examine data distribution shifts and consequent model shifts:
\vspace{-0.07in}
\label{enum:setup}
\begin{enumerate}
    \item Draw a sample of data $\mathcal{D}_1$ from the real world.
    \vspace{-0.07in}
    \item Train a binary classification model $\mathcal{M}_1$ on data $\mathcal{D}_1$.
    \vspace{-0.07in}
    \item Use a counterfactual explanation based recourse finding algorithm to generate recourses $\mathcal{CF}_1$. Recourses are found for users that are adversely classified as $-1$ by model $\mathcal{M}_1$\hideh{ from amongst the population $\mathcal{D}_1$}.
    \vspace{-0.07in}
    \item Draw a new sample of the data $\mathcal{D}_2$, that could suffer from potential distribution shifts that have occured naturally.
    \vspace{-0.07in}
    \item Train a new binary classification model $\mathcal{M}_2$ on data $\mathcal{D}_2$, using the exact same model class and hyperparameter settings as before.
    \vspace{-0.07in}
    \item Verify the predictions of model $\mathcal{M}_2$ on the recourses $\mathcal{CF}_1$. By definition, the prediction of $\mathcal{M}_1$ on these points is always $+1$, and we wish to verify that $\mathcal{M}_2$ produces the same predictions. However, if this turns out not to be the case recourses would become invalidated.
\end{enumerate}
\vspace{-0.07in}

\hideh{
The notation and steps above are summarised in figure \ref{fig:setup}.

\begin{wrapfigure}{l}{0.6\textwidth}
\vspace{-0.3in}
\centerline{\includegraphics[width=1.1\linewidth]{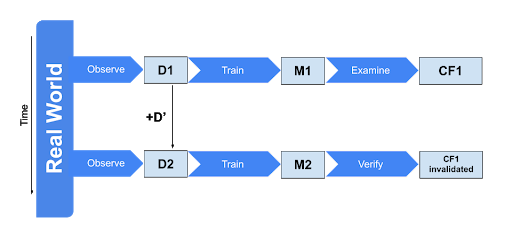}}
\vspace{-0.1in}
\caption{Problem setup: $\mathcal{D}_1$, $\mathcal{M}_1$, $\mathcal{CF}_1$, $\mathcal{D}_2$, and $\mathcal{M}_2$.}
\label{fig:setup}
\vspace{-0.1in}
\end{wrapfigure}
}

Not changing the training setup between models $\mathcal{M}_1$ and $\mathcal{M}_2$ allows us to ensure the differences in the models are entirely due to shifts in the data distribution, and not caused by hyperparameter variations or changes in the training procedures. Further, the fraction of $\mathcal{CF}_1$ predicted as $-1$ by $\mathcal{M}_2$ in step 6 above represents the ``invalidation" of recourses that we report and wish to empirically and theoretically analyze in this paper. We refer to this phenomenon as \textbf{``recourse invalidation"}, and quantify the fraction using the terms \textbf{``invalidation probability"} and \textbf{``invalidation percentage"}.
\hideh{
We can now establish the main theoretical results of this paper. We wish to show:
\vspace{-0.07in}
\begin{itemize}
    \item Cost definitions for sparse recourses are such that there exists a tradeoff between the costs and the invalidation percentages of recourses. Thus, to reduce the chances of invalidation upon updating the model, decision makers need to deliberately produce higher cost recourses than necessary.
    \vspace{-0.07in}
    \item There exists a model agnostic, non-zero lower bound on the probability of invalidation of recourses produced by sparse counterfactual based recourse generation algorithms. Thus, decision makers can never guarantee that the recourses they provide are completely invulnerable to distribution shifts.
\end{itemize}
\vspace{-0.07in}
}

\vspace{-0.05in}
\subsection{The Cost vs Invalidation Trade-off}
\vspace{-0.08in}

\begin{wrapfigure}{r}{0.5\textwidth}
\vspace{-0.35in}
\centerline{\includegraphics[width=1.1\linewidth]{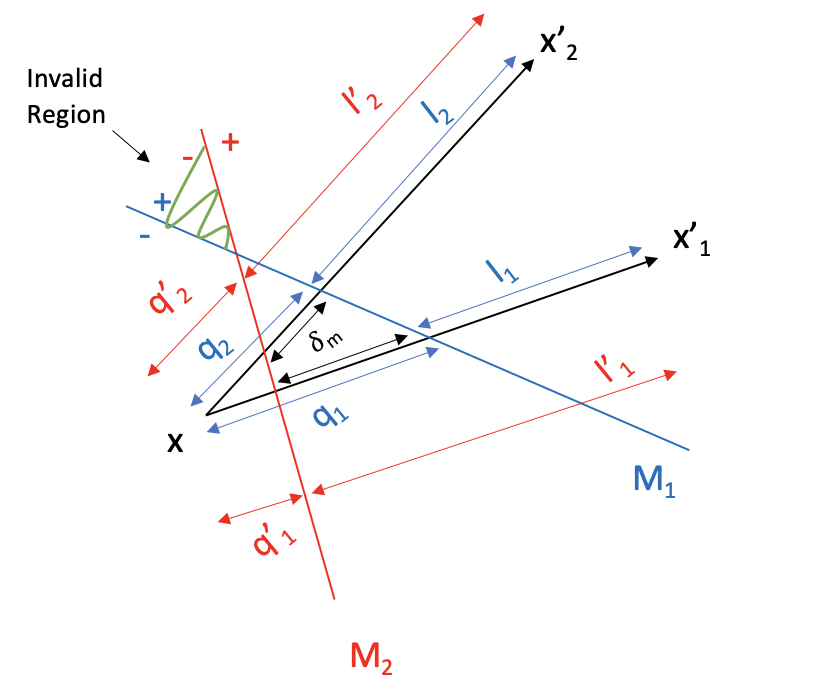}}
\vspace{-0.1in}
\caption{Abstract Diagramatic Representation of $\mathcal{M}_1$ and $\mathcal{M}_2$, with a data point $\mathcal{S}$ represented by feature-vector $x$ and two potential feature vectors $x'_1$ and $x'_2$ denoting possible recourse $\mathcal{S'}$.}
\label{fig:thm}
\vspace{-0.1in}
\end{wrapfigure}

To show that there is a trade-off between cost and invalidation percentage, we need to determine that recourses with lower costs are at high risk of invalidation, and vice-versa. Our analysis considers sparse counterfactual style recourses, presuming cost to be represented by the common Euclidean notion of distance. Our notation is illustrated in figure \ref{fig:thm} below, defining $x$ as the data point, and $x'_1$ and $x'_2$ as two potential recourses. We consider the model $\mathcal{M}_2$ to be defined as a perturbation of model $\mathcal{M}_1$ by some arbitrary magnitude $\delta_m$. Distances (measured according to the generic cost metric $d$) are measured along the vectors $x \rightarrow x'_1$ and $x \rightarrow x'_2$: $q_1$ and $q_2$ from $x$ to $\mathcal{M}_1$, and $l_1$ and $l_2$ from $\mathcal{M}_1$ to $x'_1$ and $x'_2$ respectively. We similarly have $q'_1$ and $q'_2$ from $x$ to $\mathcal{M}_2$, and $l'_1$ and $l'_2$ from $\mathcal{M}_2$ to $x'_1$ and $x'_2$ respectively. The cost function with L2 norm is very common in sparse counterfactuals, defined as $d(x,x') = ||x-x'||_2$ (\cite{wachter2018a}). Lastly, we denote the probability of invalidation for an arbitrary recourse $x'$ under model $\mathcal{M}_2$ as $Q_{x'} = \frac{1 - \mathcal{M}_2(x')}{2}$. Note that by the definition of recourse, $\mathcal{M}_1(x')=+1$ and $\mathcal{M}_2(x')=+1$ for valid recourses but $\mathcal{M}_2(x')=-1$ for invalid recourses.

\begin{theorem}
\label{thrm1}
If we have recourses $x'_1$ and $x'_2$ for a data point $x$ and model $\mathcal{M}_1$, such that $d(x,x'_1) \leq d(x,x'_1)$ then the respective expected probabilities of invalidation under model $\mathcal{M}_2$, $Q_{\mathbb{E}\left[x'_1\right]}$ and $Q_{\mathbb{E}\left[x'_2\right]}$, follow $Q_{\mathbb{E}\left[x'_1\right]} \geq Q_{\mathbb{E}\left[x'_2\right]}$.
\end{theorem}
\vspace{-0.1in}
\begin{hproof}
We know from our construction that $d(x,x'_1)=q_1+l_1$ and $d(x,x'_2)=q_2+l_2$, and also that $l'_1=l_1 \pm \delta_m$ and $l'_2=l_2 \pm \delta_m$. We assume that the small random perturbation $\delta_m$ between $\mathcal{M}_1$ and $\mathcal{M}_2$ has expected value $\mathbb{E}[\delta_m] = 0$. Further, we also assume that both $q_1$ and $q_2$ are random variables drawn from the same unknown distribution, with some arbitrary expected value $\bar{q} = \mathbb{E}\left[q_1 \right] = \mathbb{E}\left[q_2 \right]$.

Thus, referring to figure \ref{fig:thm} we get: $d(x,x'_1) \leq d(x,x'_2) \implies \mathbb{E} \left[ q_1 + l_1 \right] \leq \mathbb{E} \left[ q_2 + l_2 \right] \implies \mathbb{E} \left[l_1 \right] \pm \mathbb{E} \left[\delta_m \right] \leq \mathbb{E} \left[l_2 \right] \pm \mathbb{E} \left[\delta_m \right] \implies \mathbb{E} \left[l'_1 \right] \leq \mathbb{E} \left[l'_2 \right]$

To capture the notion that models are less confident in their predictions on points close to their decision boundaries, we construct a function $g(l') = P\left[ \mathcal{M}_2(x') = +1 \right]$, using the bijective relationship between $x'$ and $l'$. Therefore, $g$ is a monotonically increasing function, with $g(l') = -Q_{x'}$. 

We now equate the probability of invalidation $Q$ to an arbitrary function $g(l')$. The bijection between $x'$ and $l'$ allows us to define $g(l') = -Q(x')$ as a monotonic increasing function, with $g(-\infty) = 0$, $g(-\delta_m) <= 0.5$, $g(\delta_m) >= 0.5$, and $g(\infty) = 1$.  We can now apply this monotonic increasing function and continue our derivation to get: $g(\mathbb{E}\left[l'_1\right]) \leq g(\mathbb{E}\left[l'_2\right]) \implies Q_{\mathbb{E}\left[x'_1\right]} \geq Q_{\mathbb{E}\left[x'_2\right]}$. A detailed proof is included in the Appendix.
\end{hproof}

\begin{proposition}
\vspace{-0.05in}
There is a tradeoff between recourse costs (with respect to model $\mathcal{M}_1$) and recourse invalidation percentages (with respect to the updated model $\mathcal{M}_2$). Consider a hypothetical function $F$ that computes recourse costs $d(x,x')$ from expected invalidation probabilities $Q_{\mathbb{E}\left[x'\right]}$. Therefore $F\left[ Q_{\mathbb{E}\left[x'\right]} \right] = d(x,x')$ is a monotonically decreasing function: as invalidation probabilities increase (or decrease),  recourse costs decrease (or increase), and vice versa.
\end{proposition}
\vspace{-0.15in}
\begin{hproof}
Consider a hypothetical function $G = F^{-1}$ such that $G\left[ d(x,x') \right] = Q_{\mathbb{E}\left[x'\right]}$. From the proof of theorem \ref{thrm1} above we know that $d(x,x'_1) \leq d(x,x'_2) \implies Q_{\mathbb{E}\left[x'_1\right]} \geq Q_{\mathbb{E}\left[x'_2\right]}$. Thus $G$ is monotonic, and it's hypothetical inverse $F$ must also be monotonic (if it exists). A detailed proof is included in the Appendix.
\end{hproof}



\vspace{-0.05in}
For decision makers, this has key implications: cheaper costs $d(x,x'_1)$ imply higher invalidation chances $Q_{\mathbb{E}\left[ x'_1 \right]}$, and more expensive costs $d(x,x'_2)$ imply lower invalidation chances $Q_{\mathbb{E}\left[ x'_2 \right]}$. The contrapositives of these statements also hold, meaning that low invalidation probabilities imply more expensive recourse costs, and that high invalidation probabilities imply cheaper recourse costs. Therefore, we can say that \textit{there exists a tradeoff between recourse costs and invalidation probabilities.}

\hideh{
\vspace{-0.07in}
\begin{enumerate}
    \item cheaper costs $\implies$ higher invalidation
    \vspace{-0.1in}
    \item more expensive costs $\implies$ lower invalidation
    \vspace{-0.1in}
    \item lower invalidation $\implies$ more expensive costs
    \vspace{-0.1in}
    \item higher invalidation $\implies$ cheaper costs
\end{enumerate}
\vspace{-0.07in}

Taken together, these statements complete our proof and show that an increase or decrease in recourse cost must necessarily be accompanied by a decrease or increase in the probability of invalidation respectively, and vice versa. }

This establishes that recourses that have lower costs are more likely to get invalidated by the updated model $\mathcal{M}_2$. This is a critical result that demonstrates a fundamental flaw in the design of state-of-the-art recourse finding algorithms since the objective formulations in these algorithms  explicitly try to minimize these costs. By doing so, these algorithms are essentially generating recourses that are more likely to be invalidated upon model updation.

\vspace{-0.05in}
\subsection{Lower Bounds on Recourse Invalidation Probabilities}
\vspace{-0.05in}

\begin{wrapfigure}{r}{0.4\textwidth}
\vspace{-0.1in}
\centerline{\includegraphics[width=1.1\linewidth]{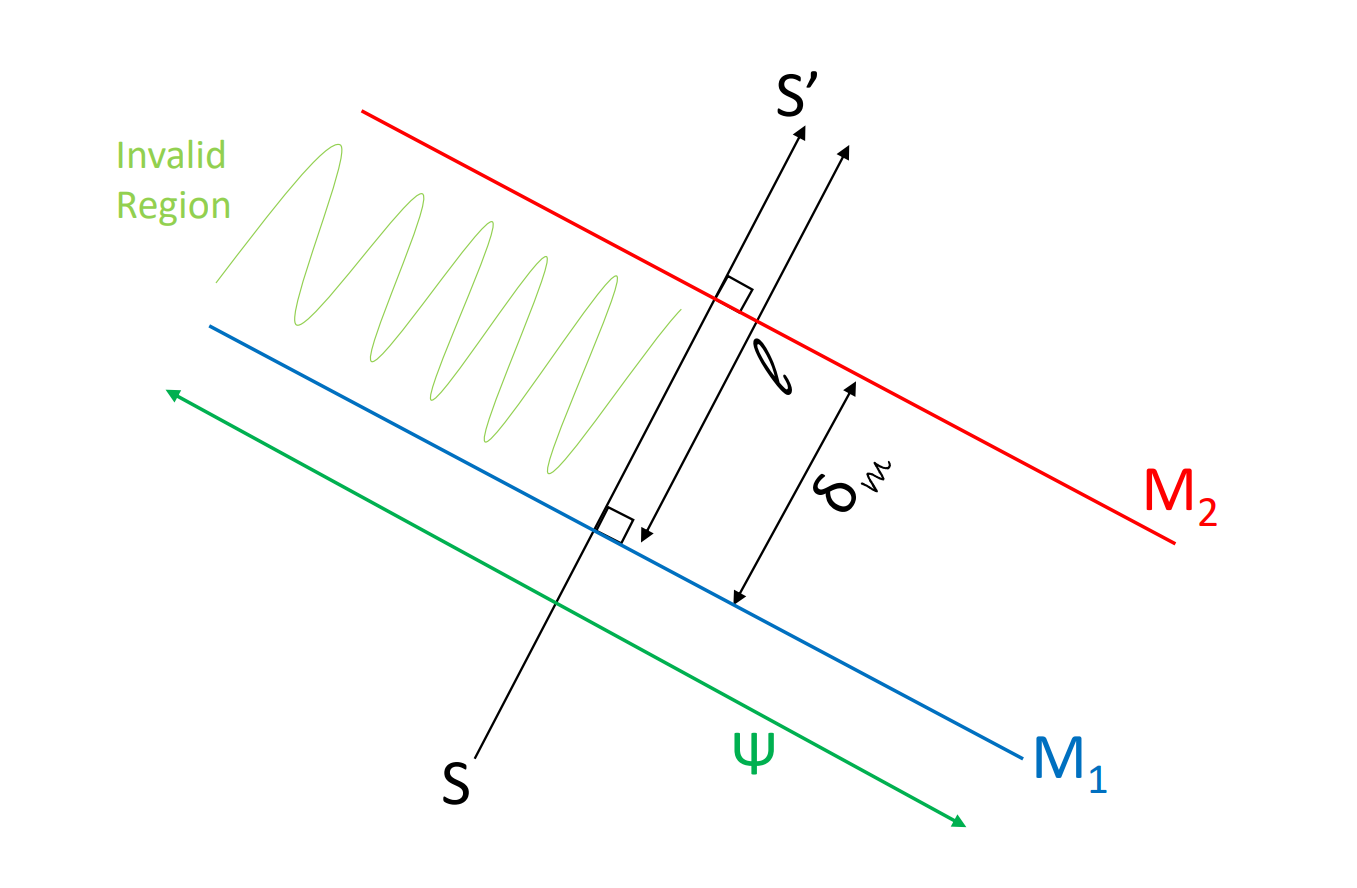}}
\vspace{-0.1in}
\caption{Parallel model perturbation of arbitrary magnitude $\delta_m$ between linear models $\mathcal{M}_1$ and $\mathcal{M}_2$. The range of the data is represented by manifold $\Psi$.}
\label{fig:thm2}
\vspace{-0.3in}
\end{wrapfigure}

In order to find general lower bounds on the probability of invalidation for recourses $\mathcal{CF}_1$ under $\mathcal{M}_2$, we first cast the couunterfactual search procedure as a Markov Decision Process, and then use this observation to hypothesize about the distributions of the recourses $\mathcal{CF}_1$. We then use these distributions to derive the lower bounds on the invalidation probability.

\begin{proposition}
The search process employed by state-of-the-art sparsity based recourse generation algorithms (e.g., \cite{wachter2018a}) is a Markov Decision Process.
\label{mdp}
\end{proposition}
\vspace{-0.15in}
\begin{hproof}
We know that the search for counterfactual explanations consists of solving the optimization problem given by $\arg \min_{\mathcal{S'}} {d(\mathcal{S'}, \mathcal{S})}$, where sparse counterfactuals are \emph{unrestricted}, and data support counterfactuals are \emph{restricted} to the set ${\mathcal{S'}: P_{data}(\mathcal{S'})>0}$. The search procedure moves through the path $\mathcal{S} \rightarrow \cdots \mathcal{S"} \cdots \rightarrow \mathcal{S'}$, looping through different possible values of $\mathcal{S"}$ until $\mathcal{M}_1(\mathcal{S"})=+1$ and cost $d(\mathcal{S},\mathcal{S"})$ is minimized - at which point the recourse $\mathcal{S'} = \mathcal{S"}$ is returned. For sparse counterfactuals, each step in this search depends only on the previous iteration, and not on the entire search path so far, that is: $P(\mathcal{S"}_{t+1} | \mathcal{S"}_t, \mathcal{S"}_{t-1} \dots \mathcal{S}) = \mathcal{S"}_{t+1} | \mathcal{S"}_t$. Thus, by satisfying this condition, the recourse generation technique is a Markov Decision Process. A detailed proof is included in the Appendix.
\end{hproof}

\begin{lemma}
If the model $\mathcal{M}_1$ is linear, then the \hideh{above proposition implies that the} distribution of recourses $\mathcal{CF}_1$ is exponential for continuous (numeric) data or geometric for categorical (ordinal) data, along the normal to the classifier hyperplane. Thus for $\mathcal{S'} \in \mathcal{CF}_1$, we have $l \sim L$ with $f(l) = \rho \cdot e^{-\rho l}$ or $f(l) = (1-\rho)^{l-1} \cdot \rho$, where $L$ is the distance from $\mathcal{S'}$ to $\mathcal{M}_1$ (figure \ref{fig:thm2}).
\label{lm:distbn}
\end{lemma}
\vspace{-0.15in}
\begin{hproof}
By Proposition \ref{mdp} above, we know that the recourse search process follows the Markov Property, and thus the distribution of the recourses must have the memoryless property. Further, since the classifier is linear, we know that an \emph{unrestricted} recourse search, such as that of sparse counterfactual based recourses, will proceed exactly along the normal to the classifier hyperplane, in the direction of increasingly positive $\mathcal{M}_1$ classification probabilities. Lastly, we assume that the probability of the counterfactual explanation search ending at any given iteration $t$ with $\mathcal{S'} = \mathcal{S"}_t$ is always constant $\rho$. Thus, for continuous data, the distribution of recourses is exponential with $\lambda = \rho$, and for ordinal data the distribution is geometric with parameter $p = \rho$. A detailed proof is included in the Appendix.
\end{hproof}

\begin{theorem}
For a given linear model $\mathcal{M}_1$ with recourses $\mathcal{CF}_1$, and a parallel linear model $\mathcal{M}_2$ with arbitrary (constant) perturbation $\delta_m$, the invalidation probabilities $Q{_\mathcal{S'}}$ are $1 - e^{-\rho \delta_m}$ for continuous (numeric) data, and $1 - (1-\rho)^{\delta_m}$ for categoric (ordinal) data.
\label{thrm3}
\end{theorem}
\vspace{-0.15in}
\begin{hproof}
Let the recourses be distributed according to some unknown arbitrary distribution with density $f(l)$, where $l \in (0, \infty)$ is the normal distance between $\mathcal{M}_1$ and $\mathcal{S'}$. Then, as is clear from the invalid region between the models $\mathcal{M}_1$ and $\mathcal{M}_2$ illustrated in figure \ref{fig:thm2}, the probability of invalidation of the recourses $\mathcal{S'}$ is $Q_{\mathcal{S'}} = \frac{1}{\Psi} \int_{\Psi} \left[ \int_0^{\delta_m} f(l) \,dl \right] \, d\psi $. Here $\psi$ is an arbitrary element of the decision boundary, within the data manifold $\Psi$.

We can now combine this result with known distributions from Lemma \ref{lm:distbn} to get: $Q_{\mathcal{S'}} = \frac{1}{\Psi} \int_{\Psi} \left[ \int_0^{\delta_m} \rho e^{-\rho l} \, dl  \right] \, d\psi = 1 - e^{-\rho \delta_m}$ for continuous features, and $Q_{\mathcal{S'}} = \frac{1}{\Psi} \int_{\Psi} \left[\sum_{l=1}^{\delta_m} (1-\rho)^{l-1}\cdot\rho \right] \, d\psi = 1 - (1-\rho)^{\delta_m}$ for ordinal features. A detailed proof is included in the Appendix. 
\end{hproof}

\hideh{
\begin{align}
    Q_{\mathcal{S'}} &= \frac{1}{\Psi} \int_{\Psi} \left[ \int_0^{\delta_m} f(l) \, dl  \right] \, d\psi \\
    &= \frac{1}{\Psi} \int_{\Psi} \left[ \int_0^{\delta_m} \rho e^{-\rho l} \, dl  \right] \, d\psi \\
    &= \frac{1}{\Psi} \int_{\Psi} \left[ 1 - e^{-\rho \delta_m}  \right] \, d\psi \\
    &= \frac{1}{\Psi} \int_{\Psi} \, d\psi \cdot \left[ 1 - e^{-\rho \delta_m}  \right] \\
    &= 1 - e^{-\rho \delta_m}
\end{align}

Similarly, for ordinal data this would be}


\begin{theorem}
For a given nonlinear model $\mathcal{M}_1$ with recourses $\mathcal{CF}_1$, and a parallel nonlinear model $\mathcal{M}_2$ with known constant perturbation $\delta_m$, the lower bound on the invalidation probabilities is achieved exactly when both models $\mathcal{M}_1$ and $\mathcal{M}_2$ are linear.
\end{theorem}
\vspace{-0.15in}
\begin{hproof}
We consider a piecewise linear approximation of the nonlinear models, with an arbitrary degree of precision. At each point in the classifier decsion boundary, we consider the piecewise linear approximation to make an angle $\theta$ with a hypothetical hyperplane in the data manifold $\Psi$. We then proceed identically as in the proof for theorem \ref{thrm3} above, with $Q_{\mathcal{S'}} = \frac{1}{\Psi \cos \theta} \int_{\Psi} \left[ \int_0^{\delta_m} f(l) \,dl \right] \, d\psi $, where the extra $\cos \theta$ term reflects that the models are locally inclined at an angle $\theta$ from the hyperplane $\psi$.

It is easy to see that $Q_{\mathcal{S'}}$ will be maximized when $\theta = 0, \forall \theta$, because $\frac{dQ_{\mathcal{S'}}}{d\theta} \propto \tan \theta = 0 \implies \theta = 0$. If each element of the piecewise linear approximation makes a constant angle of $0$ with the models, then the models themselves must be linear. Thus, the lower bound on invalidation probability for non-linear models must exactly be the invalidation probability for linear models, given the same data (manifold). A detailed proof is included in the Appendix.
\end{hproof}
\vspace{-0.1in}

Combining the last two results, for any arbitrary model $\mathcal{M}_1$, we know that another model $\mathcal{M}_2$ perturbed parallelly with magnitude would cause invalidation with the following lower bounds: $Q_{\mathcal{S'}} \geq 1 - e^{-\rho \delta_m}$, for continuous data and $Q_{\mathcal{S'}} \geq  1 - (1-\rho)^{\delta_m}$ for ordinal data.

%% file: sections/040experiments.tex
In this section we analyze recourse invalidation caused by model updates made due to naturally occurring real world distribution shifts. In our experiments we consider recourse generation methods that either directly optimise cost (sparse counterfactuals, such as \cite{wachter2018a}), promote recourses that lie in-distribution (data support counterfactuals, such as \cite{Ustun_2019}), or are generated from an assumed underlying structural causal model of the data (causal counterfactuals, such as \cite{karimi2020algorithmic1}). We also use synthetically generated datasets to perform a sensitivity analysis to further understand how the magnitude of distribution shifts affects the proportion of recourses being invalidated.

\vspace{-0.1in}
\paragraph{Setup}: We consider an initial dataset $\mathcal{D}_1$ upon which we train model $\mathcal{M}_1$, and a dataset $\mathcal{D}_2$ in which the distribution has shifted, upon which we train model $\mathcal{M}_2$. We reserve 10\% of both datasets for validation in order to perform sanity checks on the accuracies of our models $\mathcal{M}_1$ and $\mathcal{M}_2$. We then follow the experimental setup described in \ref{enum:setup} for various model types: Logistic Regression (\textbf{LR}), Random Forests (\textbf{RF}), Gradient Boosted Trees (\textbf{XGB}), Linear Support Vector Machines (\textbf{SVM}), a small Neural Network with hidden layers = [10, 10, 5] (\textbf{DNN (s)}), and a larger Neural Network with hidden layers = [20, 10, 10, 10, 5] (\textbf{DNN (l)}). The models are all treated as binary classification black-boxes with adversely classified ($\{\mathcal{S} \in \mathcal{D}_1 : \mathcal{M}_1(\mathcal{S}) = -1 \}$) people being provided recourses $\mathcal{S'} \in \mathcal{CF}_1$. Finally, we check how many of the recourses from $\mathcal{CF}_1$ are still classified adversely by the updated prediction model $\mathcal{M}2$, that is $\{\mathcal{S'} \in \mathcal{CF}_1 : \mathcal{M}_2(\mathcal{S'}) = -1 \}$. We hope that 0\% of $\mathcal{CF}_1$ has been invalidated, allowing the decision maker to keep their promise to the users when recourse was initially provided.

We conduct this entire experiment with two different recourse generation techniques: counterfactual explanations (hencefore \textbf{CFE}), from \cite{wachter2018a}; and actionable recourse (henceforth \textbf{AR}), from \cite{Ustun_2019}, details of which are included in section \ref{relwork}. To adapt CFE to non-differentiable models we use numeric differentiation and approximate gradients, and to adapt AR to non-linear models we use a local linear model approximation using LIME (\cite{ribeiro2016should}). These are standard baseline adaptations we adapt from prior works. When structural causal models are known (although this is often difficult in real world applications), we also perform experiments using \cite{karimi2020algorithmic1}. Our results containing 10-fold cross-validation accuracies for models $\mathcal{M}_1$ and $\mathcal{M}_2$ and the respective invalidation proportions are summarized in table \ref{tab:all} below. \textit{We will be releasing the code used to produce these results, including hyperparameter settings, infrastructure requirements, and runtimes, as open-source via GitHub.}

\vspace{-0.1in}
\paragraph{Datasets}: We conduct our analysis using three different real world datasets, each from a different "high-stakes" domain, to show the real world impact of our findings. The first dataset is from the domain of criminal justice (\cite{lakkaraju2016interpretable}), which contains \textit{proprietary} data from 1978 ($\mathcal{D}_1$ - 8395 points) and 1980 ($\mathcal{D}_2$ - 8595 points) respectively. This dataset contains demographic features such as race, sex, age, time-served, and employment, and a target attribute corresponding to bail decisions. This data contains an inherent \textbf{Temporal} dataset shift, as the character of the data in 1980 differs from the data in 1978. Our second dataset is from the education domain, and consists of \textit{publically available} student data collected from schools in Jordan ($\mathcal{D}_1$ - 129 points) and Kuwait ($\mathcal{D}_2$ - 122). We consider the problem of predicting grades (binary classification between pass and fail) using input predictors such as grade, holidays-taken, and class-participation (\cite{geospatial}). This data contains an inherent \textbf{Geospatial} distribution shift as the character of student data varies across countries. Lastly, from the domain of finance, we consider the \textit{publically available} German credit dataset ($\mathcal{D}_1$ - 900 points) (\cite{UCI}), along with its updated version ($\mathcal{D}_2$ - 900 points) (\cite{ngcredit}). This This is a credit scoring problem using features such as the applicants loan amount, employment history, and age as predictors. The data here represents a \textbf{Data Correction} based distribution shift, as the character of the data can be said to change due to a change in the data preprocessing step. Unlike the previous datasets, in this case we also have access to an underlying causal model from \cite{karimi2020algorithmic1}, and we are thus able to additionally perform experiments using the causal recourse generation algorithm proposed here.

\begin{table}[!htp]\centering
\scriptsize
\begin{tabular}{lc|rcr|rcr|rcr}\toprule
& & \multicolumn{3}{c}{Temporal Shift} & \multicolumn{3}{c}{Geospatial Shift} & \multicolumn{3}{c}{Data Correction Shift} \\
\textbf{Algorithm} &\textbf{Model} &\textbf{M1 acc} &\textbf{M2 acc} &\textbf{Inv. \%} &\textbf{M1 acc} &\textbf{M2 acc} &\textbf{Inv. \%} &\textbf{M1 acc} &\textbf{M2 acc} &\textbf{Inv. \%} \\ \midrule \midrule
\multirow{6}{*}{\textbf{AR}} &LR &94 &95.4 &96.6 &88 &93 &76.6 &71 &75 &7.79 \\
&RF &99 &99.5 &0.05 &89 &92 &\textit{NAN} &73 &73 &35 \\
&XGB &100 &99.7 &0 &85 &93 &\textit{NAN} &74 &75 &8 \\
&SVM &81 &78.9 &3.05 &80 &91 &90 &63 &69 &100 \\
&DNN (s) &99 &99.4 &19.26 &83 &87 &\textit{NAN} &68 &69 &\textit{NAN} \\
&DNN (l) &99 &99.6 &0 &82 &93 &\textit{NAN} &66 &67 &0 \\
\midrule
\multirow{6}{*}{\textbf{CFE}} &LR &94 &95.4 &98.29 &88 &93 &65.96 &71 &75 &3.9 \\
&RF &99 &99.5 &0.71 &89 &92 &76.47 &73 &73 &36.82 \\
&XGB &100 &99.7 &0.46 &85 &93 &57.14 &74 &75 &23.72 \\
&SVM &81 &78.9 &100 &80 &91 &100 &63 &69 &0 \\
&DNN (s) &99 &99.4 &91.38 &83 &87  &50 &68 &69 &\textit{NAN} \\
&DNN (l) &99 &99.6 &0.13 &82 &93 &30.3 &66 &67 &0 \\
\midrule
\multirow{6}{*}{\textbf{Causal}} &LR & & & & & & &69 &71 &0 \\
&RF & & \centering unknown & & & \centering unknown & &65 &64 &96.09 \\
&XGB & & \centering causal & & & \centering causal & &64 &68 &12.5 \\
&DNN (s) & & \centering model & & & \centering model & &69 &70 &\textit{NAN} \\
&DNN (l) & & & & & & &69 &70 &\textit{NAN} \\
\bottomrule
\vspace{0.01in}
\end{tabular}
\caption{ Recourse Invalidation Proportions caused by Model Updates due to various (Temporal, Geospatioal, or Data Correction) Distribution Shifts. }
\label{tab:all}
\end{table}

\vspace{-0.1in}
\paragraph{Temporal Distribution Shifts}: Columns 3, 4, and 5 showcase the results on the bail dataset from the criminal justice domain, which contains a temporal distribution shift. We can see that most models have high (> 95\%) accuracy, and yet the invalidation proportions vary significantly between AR and CFE, however this trend does not seem to be corellated with model accuracies. While there are some cases with no invalidation, the CFE algorithm, when applied to an SVM model, creates a situation where all of the recourses generated end up getting invalidated. This clearly demonstrates the potential harm of recourse invalidation. We are unable to perform experiments using a causal recourse generation technique because the underlying causal distribution is unknown for this dataset.

\vspace{-0.1in}
\paragraph{Geospatial Distribution Shifts}: Columns 6, 7, and 8 showcase results on the schools dataset from the education domain, which contains a geospatial data distribution shift. We see a minimum invalidation of 30\% on this dataset, indicating there are no good situations for a decision maker to provide algorithmic recourses in the scenario. We see multiple \textit{NAN} values for recourse invalidation for the AR method on this dataset. These represent those situations when no recourses could be generated (that is, $\mathcal{CF}_1$ is empty, and thus the proportion of recourses from $\mathcal{CF}_1$ that are invalidated is undefined). We also see here that even though model accuracies are increasing, recourse invalidation remains an observable issue, indicating that this phenomenon cannot be explained away as a modelling error made by decision makers, but instead is inherent to distribution shifts. The lack of an underlying structural causal model again precludes us from being able to conduct experiments using causal recourse generation techniques on this data.

\vspace{-0.1in}
\paragraph{Data Correction Distribution Shifts}: Columns 9, 10, and 11 showcase results on the German credit datasets, which contain a data-correction related distribution shifts. Often decision makers might initially deploy a model trained on inaccurate or corrupt data, or data that suffers from selection biases. As the decision makers improve their training datasets and redeploy their models, a distribution shift would occur, which is captured by this experiment. Again, \textit{NAN} values indicate that no recourses were found, and thus the proportion of those invalidated is undefined. Here, we see extreme behaviour - SVM shows 100\% invalidation with AR, but 0\% with CFE. On the other hand, XGB has higher invalidation with CFE than with AR. This further demonstrates that the phenomenon of recourse invalidation is independent of model types or specific recourse generation algorithms. Lastly, we see that even causally generated recourses are vulnerable to the problem of recourse invalidation. Interestingly, the RF and XGB models are the worst affected, even though the temporal distribution shift results might have led us to believe these were the most robust models. Thus, it appears that the advantages (\cite{karimi2020algorithmic2, venkat, Barocas_2020}) of causally generated recourses over other recourse generation algorithms do not protect us from recourse invalidation due to model updates caused by distribution shifts.

\vspace{-0.1in}
\paragraph{Sensitivity Analysis}: We have demonstrated empirically that distribution shifts cause recourse invalidation, and we now wish to qualify whether larger distribution shifts lead to greater recourse invalidation. To do this we construct synthetic datasets and precisely control the type of distribution shift. We start with a fixed distribution $\mathcal{D}_1$, which has two independent predictors $X0$ and $X1$, both drawn from a standard normal distribution, with the binary target attribute defined linearly as $Y = (X0 + X1 \geq 0)$. We then train a logistic regression model $\mathcal{M}_1$ and generate recourses $\mathcal{CF}_1$ from either AR or CFE. Finally, we shift the distribution $\mathcal{D}_2$ according to some shift parameter $\alpha$ and construct logistic regression model $\mathcal{M}_2$, and analyse the relation between recourse invalidation percentage and $\alpha$. We consider two scenarios, with a shift parameter $\alpha$ defining amount of shift between $\mathcal{D}_2$ and $\mathcal{D}_1$:

\vspace{-0.05in}
\begin{enumerate}
    \item \textbf{Shifting target}: where the predictor distribution ($X0$ and $X1$) stays constant, but the $\mathcal{D}_2$ target attribute is defined as $Y = (X0 + \alpha X1 \geq 0)$, for a \% shift of $\alpha \in (-60\%, 60\%)$.
    \item \textbf{Shifting predictors}: where the definition of the target stays constant ($Y = (X0 + X1 \geq 0)$), but we shift the mean of the predictor distribution in $\mathcal{D}_2$ from $(X0, X1) = (0, 0)$ to $(X0, X1) = (\alpha, \alpha)$.
\end{enumerate}
\vspace{-0.03in}

\begin{figure}
\begin{subfigure}{.49\textwidth}
  \includegraphics[width=1.0\linewidth]{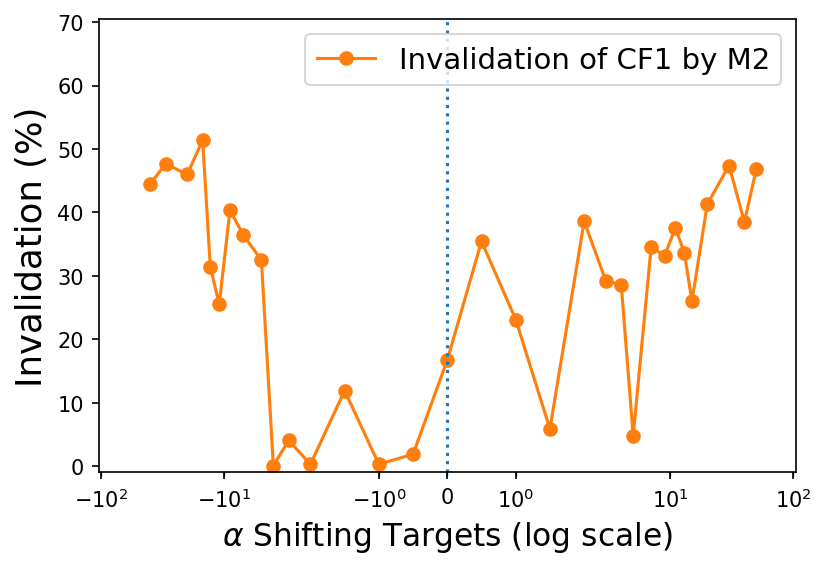}  
  \caption{Recourse Invalidation vs Drifting Targets [AR]}
  \label{fig:sens1}
    \end{subfigure}
\begin{subfigure}{.49\textwidth}
  \includegraphics[width=1.0\linewidth]{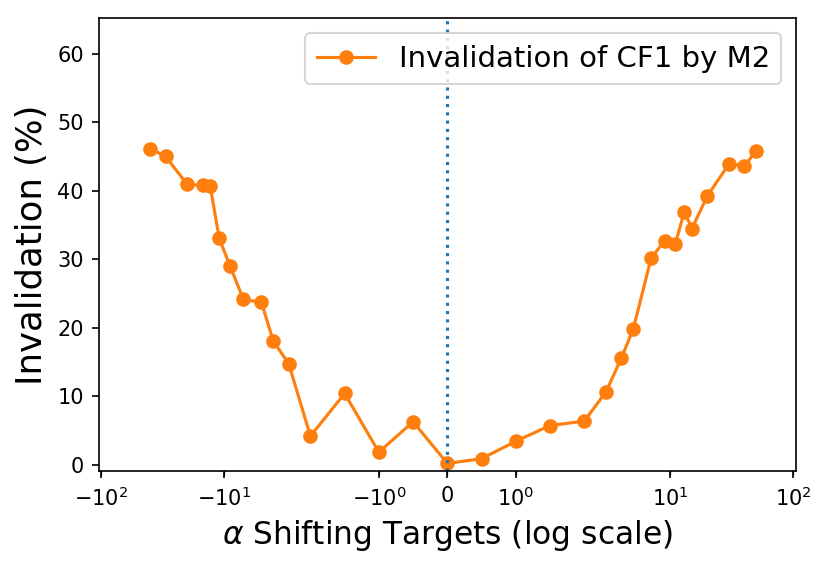}  
  \caption{Recourse Invalidation vs Drifting Targets [CFE]}
  \label{fig:sens2}
\end{subfigure}
\newline
\begin{subfigure}{.49\textwidth}
  \includegraphics[width=1.0\linewidth]{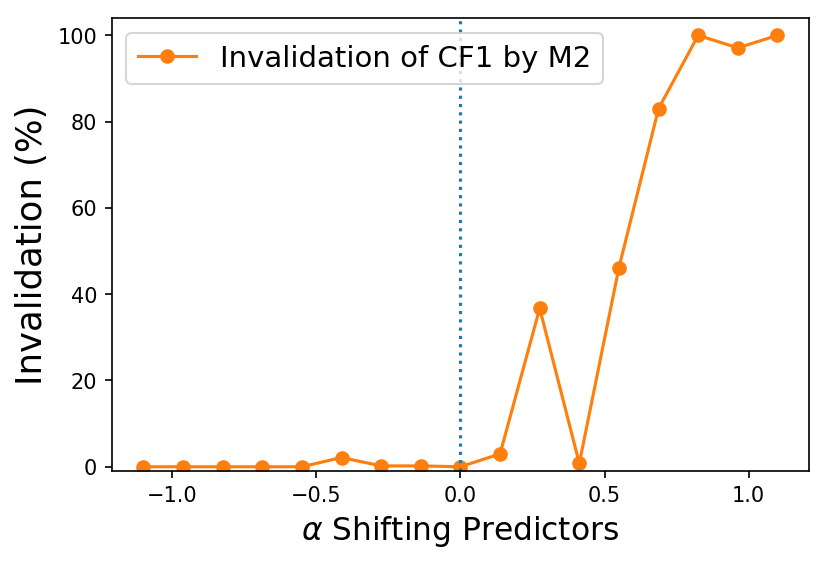}  
  \caption{Recourse Invalidation vs Drifting Predictors [AR]}
  \label{fig:sens3}
\end{subfigure}
\begin{subfigure}{.49\textwidth}
  \includegraphics[width=1.0\linewidth]{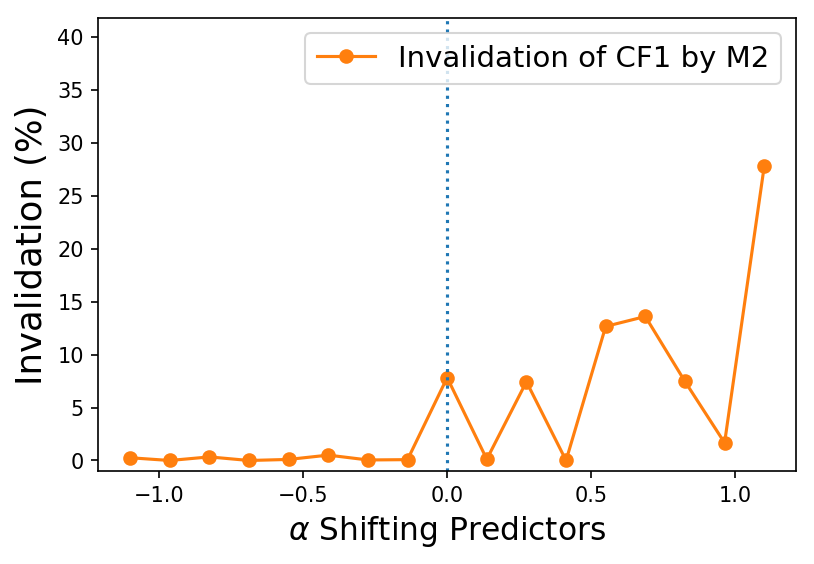}  
  \caption{Recourse Invalidation vs Drifting Predictors [CFE]}
  \label{fig:sens4}
\end{subfigure}
\vspace{0.03in}
\caption{Sensitivity Analysis: Recourse Invalidation Increases with Increasing Drift. (All models have accuracy > 99\%)}
\vspace{-0.01in}
\label{fig:sens_all}
\end{figure}

As figures \ref{fig:sens1} through \ref{fig:sens4} demonstrate, recourse invalidation increases with increasing distribution shift, but this is not universal. There may be some distribution shifts that the recourses generated are robust to, and others that result in upto 100\% of generated recourses getting invalidated. Particularly, when $\alpha$ is negative while shifting predictors (with this particular target distribution), recourses generated are robust. In all other cases there is significant invalidation that increases with increasing $\alpha$. It is also noteworthy that even with data support based counterfactual recourse AR, we do not uniformly see less invalidation, as postulated by \cite{Ustun_2019}. Sometimes the recourses generated are indeed more robust, but they also often fail miserably with up to 100\% invalidation. This would suggest that data support counterfactual based recourses are actually less predictable than sparse counterfactual based recourses, while still being vulnerable to invalidation, and could thus potentially be more risky for real world decision makers.

\hideh{
\begin{figure}
\begin{subfigure}{.24\textwidth}
  \centering
  \includegraphics[width=1.0\linewidth]{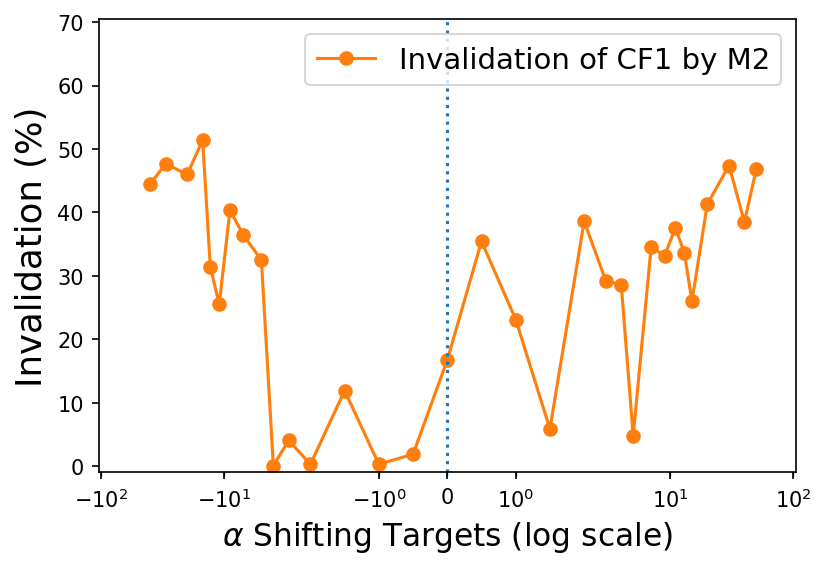}  
  \caption{Recourse Invalidation vs Drifting Targets [AR]}
  \label{fig:sens1}
    \end{subfigure}
\begin{subfigure}{.24\textwidth}
  \centering
  \includegraphics[width=1.0\linewidth]{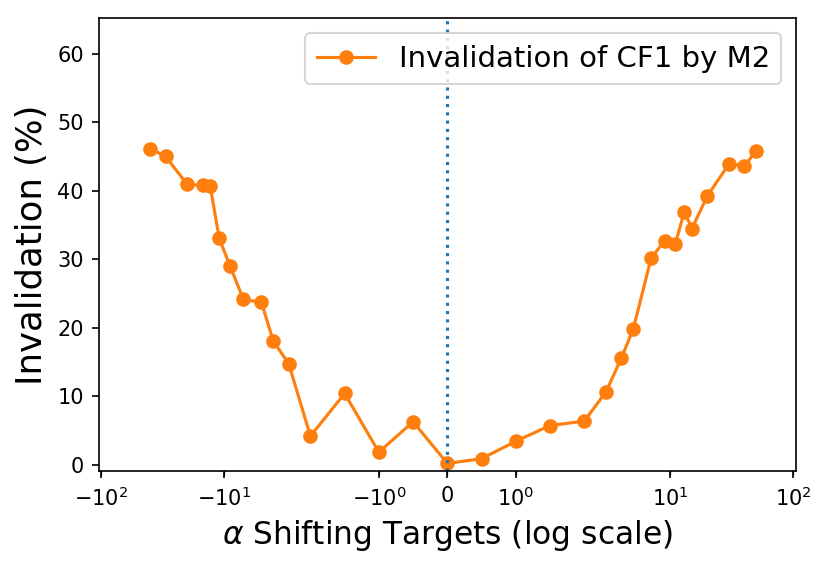}  
  \caption{Recourse Invalidation vs Drifting Targets [CFE]}
  \label{fig:sens2}
\end{subfigure}
\begin{subfigure}{.24\textwidth}
  \centering
  \includegraphics[width=1.0\linewidth]{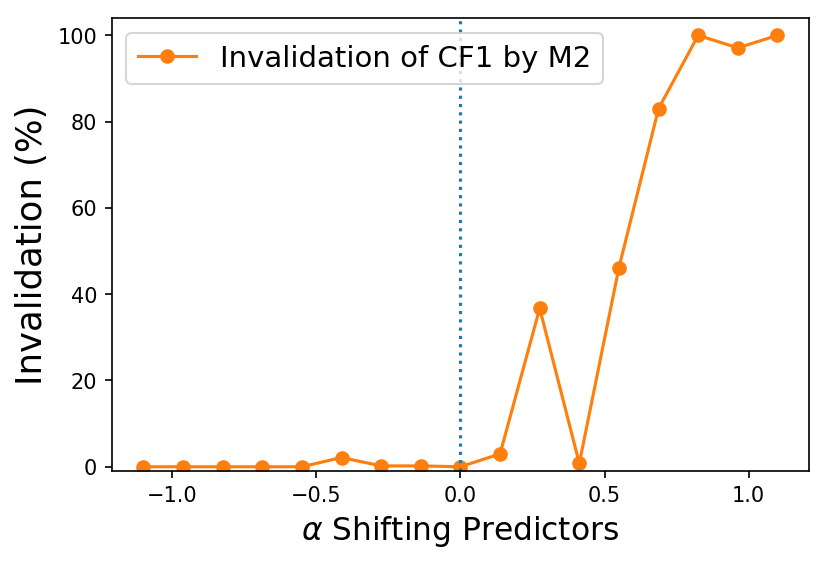}  
  \caption{Recourse Invalidation vs Drifting Predictors [AR]}
  \label{fig:sens3}
\end{subfigure}
\begin{subfigure}{.24\textwidth}
  \centering
  \includegraphics[width=1.0\linewidth]{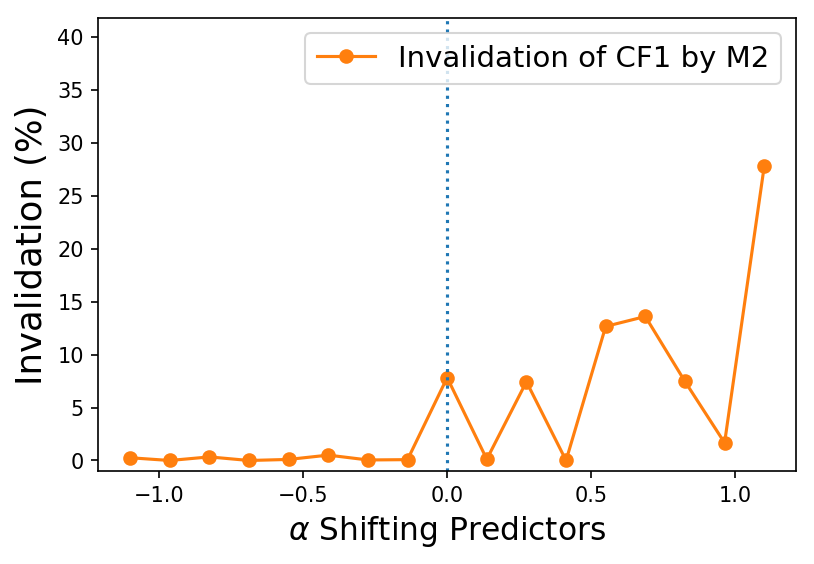}  
  \caption{Recourse Invalidation vs Drifting Predictors [CFE]}
  \label{fig:sens4}
\end{subfigure}
\caption{Sensitivity Analysis: Recourse Invalidation Increases with Increasing amount of Drift. (All models have accuracy > 99\%.)}
\label{fig:sens_all}
\end{figure}
}

%% file: sections/050conclusion.tex
In this paper, we analyse the impact of distribution shifts on recourses generated by state-of-the-art algorithms. We conduct multiple real world experiments and show that distribution shifts can cause significant invalidation of generated recourses, jeopardizing trust in decision makers. This is contrary to the goals of a decision maker providing recourse to their users. We show theoretically that there is a trade-off between minimising cost and providing robust, hard-to-invalidate recourses when using sparse counterfactual generation techniques. We also find the lower bounds on the probability of invalidation of recourses when model updates are of known perturbation magnitude. While our theory pertains only to sparse counterfactual based recourses, we demonstrate experimentally that invalidation problems due to distribution shift persist in practice not only with data support based counterfactual recourse, but also with causally generated recourses. Theoretical examination of recourse invalidation when using these recourse generation methods remains as future work.

This work paves the way for several other interesting future directions too. It would be interesting to develop novel recourse finding strategies that do not suffer from the drawbacks of existing techniques and are robust to distribution shifts. This could involve iteratively finding and examining recourses as part of the model training phase, in order to ensure that recourses with high likelihood of invalidation are never prescribed after model deployment.

\hideh{ We investigate whether this phenomenon is an artifact of insufficient training of the initial model, but find this to be untrue. We motivate theoretically that the cost functions that are defined and minimised by modern recourse finding algorithms incentivise the generated recourses to be at high risk of invalidation, and therefore fragile and susceptible to invalidation by distribution shifts. 

Thus, the problem of distribution shifts invalidating recourses and counterfactual explanations seems to be a direct result of current recourse finding technologies, rather than of the properties of the initial model. We also observe this phenomenon empirically, showing that recourses that are close to the decision boundary are not only preferred by recourse finding mechanisms, but are also more likely to be invalidated upon model updation than those recourses that are far from the decision boundary.

This work paves way for several interesting future directions. First, it would be interesting to develop novel recourse finding strategies that do not suffer from the drawbacks of existing techniques and are robust to distribution shifts. This could involve iteratively finding and examining recourses as part of the model training phase, in order to ensure that recourses with high likelihood of invalidation are never prescribed after deployment. 
}

%% file: sections/060appendix.tex
\section{Theoretical Analysis}

\subsection{Preliminaries}

We consider a black box binary classifier $\mathcal{M}$ trained on a dataset of real world users $\mathcal{D}$ and for each user $\mathcal{S}$ in $\mathcal{D}$, let us consider the corresponding recourse found by a recourse generation algorithm $Alg$ to be $\mathcal{S'}$.

Further, let the set of adversely (negatively) classified users be $\mathcal{D}^{M-}$, positively classified users be $\mathcal{D}^{M+}$, and the set of all recourses be $\mathcal{CF}_1$. Our definitions can be thus written as follows:


\begin{align}
 \mathcal{D}^{M+} &= \{\mathcal{S} \in \mathcal{D} : \mathcal{M}(\mathcal{S}) = +1\} \\
 \mathcal{D}^{M-} &= \{\mathcal{S} \in \mathcal{D} : \mathcal{M}(\mathcal{S}) = -1\} \\
 \mathcal{CF}_1 &= \{ \mathcal{S} \in \mathcal{D}^{M-}, \mathcal{S'} : Alg(\mathcal{S}) = \mathcal{S'} \} 
\end{align}

From our construction the following three statements always hold:
\begin{align}
\mathcal{M}(\mathcal{S}) &= -1 \;\; \forall  \;\; \mathcal{S} \in \mathcal{D}^{M-} \\
\mathcal{M}(\mathcal{S}) &= +1 \;\; \forall  \;\; \mathcal{S} \in \mathcal{D}^{M+} \\
\mathcal{M}(\mathcal{S'}) &= +1 \;\; \forall  \;\; \mathcal{S'} \in \mathcal{CF}_1 
\end{align}

All recourse generation methods essentially perform a search starting from a given data point $\mathcal{S}$, going towards $\mathcal{S'}$, by repeatedly polling the input-space of the black box model $\mathcal{M}$ along the path $\mathcal{S} \rightarrow \cdots \mathcal{S"} \cdots \rightarrow \mathcal{S'}$. There are many ways in which to guide this path when searching for counterfactual explanations. Sparse recourse generation techniques (\cite{Pawelczyk_2020}) operate on a given data point $\mathcal{S}$, and an arbitrary cost function $d(\mathcal{S}, \mathcal{S'})$ (usually set to $d(\mathcal{S}, \mathcal{S'}) = ||\mathcal{S}-\mathcal{S'}||_p$, $p \in \{1, 2\}$). For example, \cite{wachter2018a} define optimal recourses according to eqn \ref{app:eqn:wachter}. Data support recourses restrict the counterfactual search to those $\mathcal{S'}$ that are found in the data distribution (with high probability), as in eqn \ref{app:eqn:berk}. \cite{Ustun_2019} use a heuristic to try and optimise this using linear programming tools. \label{app:sec:prelims}


\begin{align}
    \mathcal{S'} &= \arg \min_{\mathcal{S'}} \, {d(\mathcal{S}, \mathcal{S'})} \label{app:eqn:wachter} \\
    \mathcal{S'} &= \arg \min_{\mathcal{S'}: P_{data}(\mathcal{S'})>0} \, {d(\mathcal{S}, \mathcal{S'})} \label{app:eqn:berk}
\end{align}

\subsection{The Cost vs Invalidation Trade-off}
To show that there is a trade-off between cost and invalidation percentage, we need to determine that recourses with lower costs are at high risk of invalidation, and vice-versa. Our analysis considers sparse counterfactual style recourses \cite{Pawelczyk_2020}, presuming cost to be represented by the common Euclidean notion of distance. Our notation is illustrated in figure \ref{app:fig:thm} below, defining $x$ as the data point, and $x'_1$ and $x'_2$ as two potential recourses. We consider the model $\mathcal{M}_2$ to be defined as a perturbation of model $\mathcal{M}_1$ by some arbitrary magnitude $\delta_m$. Distances (measured according to the generic cost metric $d$) are measured along the vectors $x \rightarrow x'_1$ and $x \rightarrow x'_2$: $q_1$ and $q_2$ from $x$ to $\mathcal{M}_1$, and $l_1$ and $l_2$ from $\mathcal{M}_1$ to $x'_1$ and $x'_2$ respectively. We similarly have $q'_1$ and $q'_2$ from $x$ to $\mathcal{M}_2$, and $l'_1$ and $l'_2$ from $\mathcal{M}_2$ to $x'_1$ and $x'_2$ respectively. The cost function with L2 norm is very common in sparse counterfactuals, defined as $d(x,x') = ||x-x'||_2$\cite{wachter2018a}. Lastly, we denote the probability of invalidation for an arbitrary recourse $x'$ under model $\mathcal{M}_2$ as $Q_{x'} = \frac{1 - \mathcal{M}_2(x')}{2}$. Note that by the definition of recourse, $\mathcal{M}_1(x')=+1$ and $\mathcal{M}_2(x')=+1$ for valid recourses but $\mathcal{M}_2(x')=-1$ for invalid recourses.

\begin{figure}[ht]
\begin{center}
\centerline{\includegraphics[width=0.7\linewidth]{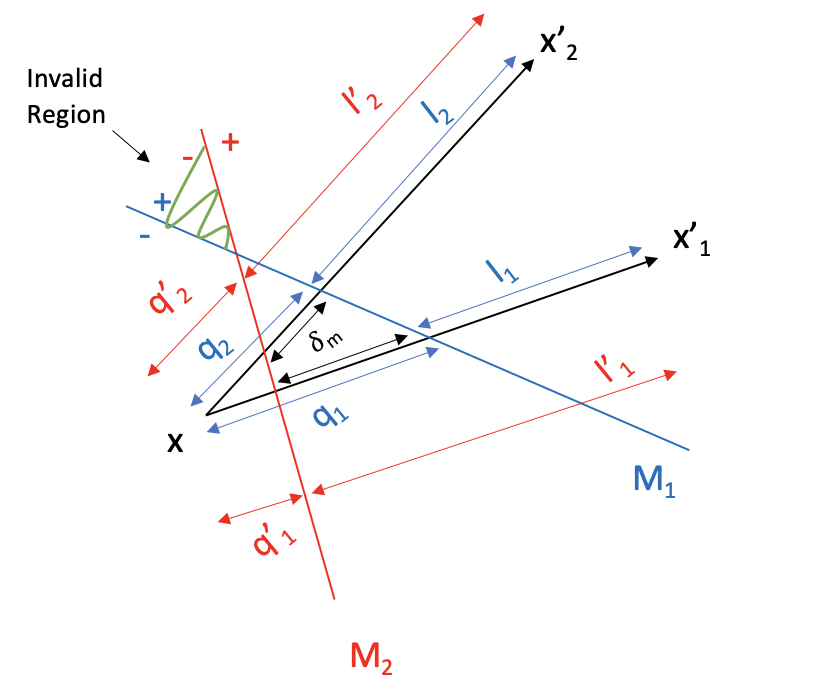}}
\caption{Abstract Diagramatic Representation of $\mathcal{M}_1$ and $\mathcal{M}_2$, with a data point $\mathcal{S}$ represented by feature-vector $x$ and two potential feature vectors $x'_1$ and $x'_2$ denoting possible recourse $\mathcal{S'}$.}
\label{app:fig:thm}
\end{center}
\end{figure}

\begin{theorem}
\label{app:thrm1}
If we have recourses $x'_1$ and $x'_2$ for a data point $x$ and model $\mathcal{M}_1$, such that $d(x,x'_1) \leq d(x,x'_1)$ then the respective expected probabilities of invalidation under model $\mathcal{M}_2$, $Q_{\mathbb{E}\left[x'_1\right]}$ and $Q_{\mathbb{E}\left[x'_2\right]}$, follow $Q_{\mathbb{E}\left[x'_1\right]} \geq Q_{\mathbb{E}\left[x'_2\right]}$.
\end{theorem}
\begin{proof}
We know from our construction that $d(x,x'_1)=q_1+l_1$ and $d(x,x'_2)=q_2+l_2$, and also that $l'_1=l_1 \pm \delta_m$ and $l'_2=l_2 \pm \delta_m$. We assume that the small random perturbation $\delta_m$ between $\mathcal{M}_1$ and $\mathcal{M}_2$ has expected value $\mathbb{E}[\delta_m] = 0$. Further, we also assume that both $q_1$ and $q_2$ are random variables drawn from the same unknown distribution, with some arbitrary expected value $\bar{q} = \mathbb{E}\left[q_1 \right] = \mathbb{E}\left[q_2 \right]$.

\begin{align}
    d(x,x'_1) &\leq d(x,x'_2) \\
    q_1 + l_1 &\leq q_2 + l_2 \\
    \mathbb{E} \left[ q_1 + l_1 \right] &\leq \mathbb{E} \left[ q_2 + l_2 \right] \\
    \mathbb{E} \left[ q_1 \right] + \mathbb{E} \left[ l_1 \right] &\leq \mathbb{E} \left[ q_2 \right] + \mathbb{E} \left[ l_2 \right] \\
    \bar{q} + \mathbb{E} \left[l_1 \right] &\leq \bar{q} + \mathbb{E} \left[l_2 \right] \\
    \mathbb{E} \left[l_1 \right] &\leq \mathbb{E} \left[l_2 \right] 
    \end{align}
\begin{align}
    \mathbb{E} \left[l_1 \right] \pm 0 &\leq \mathbb{E} \left[l_2 \right] \pm 0 \\
    \mathbb{E} \left[l_1 \right] \pm \mathbb{E} \left[\delta_m \right] &\leq \mathbb{E} \left[l_2 \right] \pm \mathbb{E} \left[\delta_m \right] \\
    \mathbb{E} \left[l_1 \pm \delta_m \right] &\leq \mathbb{E} \left[l_2 \pm \delta_m \right] \\
    \mathbb{E} \left[l'_1 \right] &\leq \mathbb{E} \left[l'_2 \right]
\end{align}

To capture the notion that models are less confident in their predictions on points close to their decision boundaries, we construct a function $g(l') = P\left[ \mathcal{M}_2(x') = +1 \right]$, using the bijective relationship between $x'$ and $l'$. Therefore, $g$ is a monotonically increasing function, with $g(l') = -Q_{x'}$. 

We now equate the probability of invalidation $Q$ to an arbitrary function $g(l')$. The bijection between $x'$ and $l'$ allows us to define $g(l') = -Q(x')$ as a monotonic increasing function, with $g(-\infty) = 0$, $g(-\delta_m) <= 0.5$, $g(\delta_m) >= 0.5$, and $g(\infty) = 1$. Lastly, we can now apply this monotonic increasing function and continue our derivation as follows:

\begin{align*}
    g(\mathbb{E}\left[l'_1\right]) &\leq g(\mathbb{E}\left[l'_2\right]) \\
    -Q_{\mathbb{E}\left[x'_1\right]} &\leq -Q_{\mathbb{E}\left[x'_2\right]} \\
    Q_{\mathbb{E}\left[x'_1\right]} &\geq Q_{\mathbb{E}\left[x'_2\right]}
\end{align*}
which gives us the intended result.
\end{proof}

\begin{proposition}
There is a tradeoff between recourse costs (with respect to model $\mathcal{M}_1$) and recourse invalidation percentages (with respect to the updated model $\mathcal{M}_2$). Consider a hypothetical function $F$ that computes recourse costs $d(x,x')$ from expected invalidation probabilities $Q_{\mathbb{E}\left[x'\right]}$. Therefore $F\left[ Q_{\mathbb{E}\left[x'\right]} \right] = d(x,x')$ is a monotonically decreasing function: as invalidation probabilities increase (or decreases), recourse costs decrease (or increase) and vice versa.
\end{proposition}
\begin{proof}
Consider a hypothetical function $G = F^{-1}$ such that $G\left[ d(x,x') \right] = Q_{\mathbb{E}\left[x'\right]}$. From the proof of theorem \ref{app:thrm1} above we know that $d(x,x'_1) \leq d(x,x'_2) \implies Q_{\mathbb{E}\left[x'_1\right]} \geq Q_{\mathbb{E}\left[x'_2\right]}$. Thus $G$ is monotonic, and it's hypothetical inverse $F$ must also be monotonic (if it exists). This implies: \textbf{(1)} cheaper costs $d(x,x'_1)$ imply higher invalidation chances $Q_{\mathbb{E}\left[ x'_1 \right]}$, and also \textbf{(2)} that more expensive costs $d(x,x'_2)$ imply lower invalidation chances $Q_{\mathbb{E}\left[ x'_2 \right]}$. We can also take contrapositives of these first two statements, giving us the third and fourth statement:

\begin{enumerate}
    \item cheaper costs $\implies$ higher invalidation
    \item more expensive costs $\implies$ lower invalidation
    \item lower invalidation $\implies$ more expensive costs
    \item higher invalidation $\implies$ cheaper costs
\end{enumerate}

Taken together, these statements complete our proof and show that an increase (or decrease) in recourse invalidation probabilities must necessarily be accompanied by a decrease (or increase) in recourse costs respectively, and vice versa. Therefore, we can say that \textit{there exists a tradeoff between recourse costs and invalidation probabilities.}
\end{proof}


\subsection{Lower Bounds on Recourse Invalidation Probabilities}

In order to find general lower bounds on the probability of invalidation for recourses $\mathcal{CF}_1$ under $\mathcal{M}_2$, we first cast the counterfactual search procedure as a Markov Decision Process, and then use this observation to hypothesize about the distributions of the recourses $\mathcal{CF}_1$. We then use these distributions to derive the lower bounds on the invalidation probability.

\begin{proposition}
The search process employed by state-of-the-art sparsity based recourse generation algorithms (e.g., \cite{wachter2018a}) satisfies the Markovian property and is a Markov Decision Process.
\label{app:mdp}
\end{proposition}
\begin{proof}
We know that the search for counterfactual explanations consists of solving the optimization problem given by:
\begin{equation}
    \arg \min_{\mathcal{S'}} {d(\mathcal{S'}, \mathcal{S})}
\end{equation}
where sparse counterfactuals are \emph{unrestricted}, and data support counterfactuals are \emph{restricted} to the set $\{\mathcal{S'}: P_{data}(\mathcal{S'})>0\}$. 

The search procedure moves through the path $\mathcal{S} \rightarrow \cdots \mathcal{S"} \cdots \rightarrow \mathcal{S'}$, looping through different possible values of $\mathcal{S"}$ until it reaches some $\mathcal{S'}$ where $\mathcal{M}_1(\mathcal{S'})=+1$ and cost $d(\mathcal{S},\mathcal{S'})$ is minimized and then that recourse $\mathcal{S'}$ is returned. 

In case of methods generating sparse counterfactuals, each step in the search depends only on the previous iteration, and not on the entire search path so far, that is: $P(\mathcal{S"}_{t+1} | \mathcal{S"}_t, \mathcal{S"}_{t-1} \dots \mathcal{S}) = P(\mathcal{S"}_{t+1} | \mathcal{S"}_t)$. Therefore, the search process employed by state-of-the-art sparsity based recourse generation algorithms satisfies the Markovian property. 

Furthermore, the search process can also be modeled as a Markov Decision Process (MDP) represented by the 4-tuple $(S, A, P, R)$. Each possible counterfactual corresponds a state in the state space $S$ with the initial state being the original instance for which counterfactual must be found. The terminal states correspond to all those counterfactuals for which the model prediction turns out to be positive i.e., $\mathcal{M}_1(\mathcal{S'})=+1$. The action set $A$ constitutes the changes that need to be made to go from one possible counterfactual to another (e.g., "age + 2 years").  Each action from any given state unambiguously and deterministically leads to a single new state. So, the transition probabilities in $P$ are either $0$ or $1$. Lastly, the reward function $R$ is defined as follows:  If an action $a \in A$ from some state $s \in S$ leads to a terminal state (i.e., to a counterfactual for which model prediction is positive), then the immediate reward is $1$, otherwise the immediate reward is $0$. 

Thus, the search process for sparsity based recourse generation techniques can be modeled as a Markov Decision Process.
\end{proof}

\begin{figure}[ht]
\begin{center}
\centerline{\includegraphics[width=0.7\linewidth]{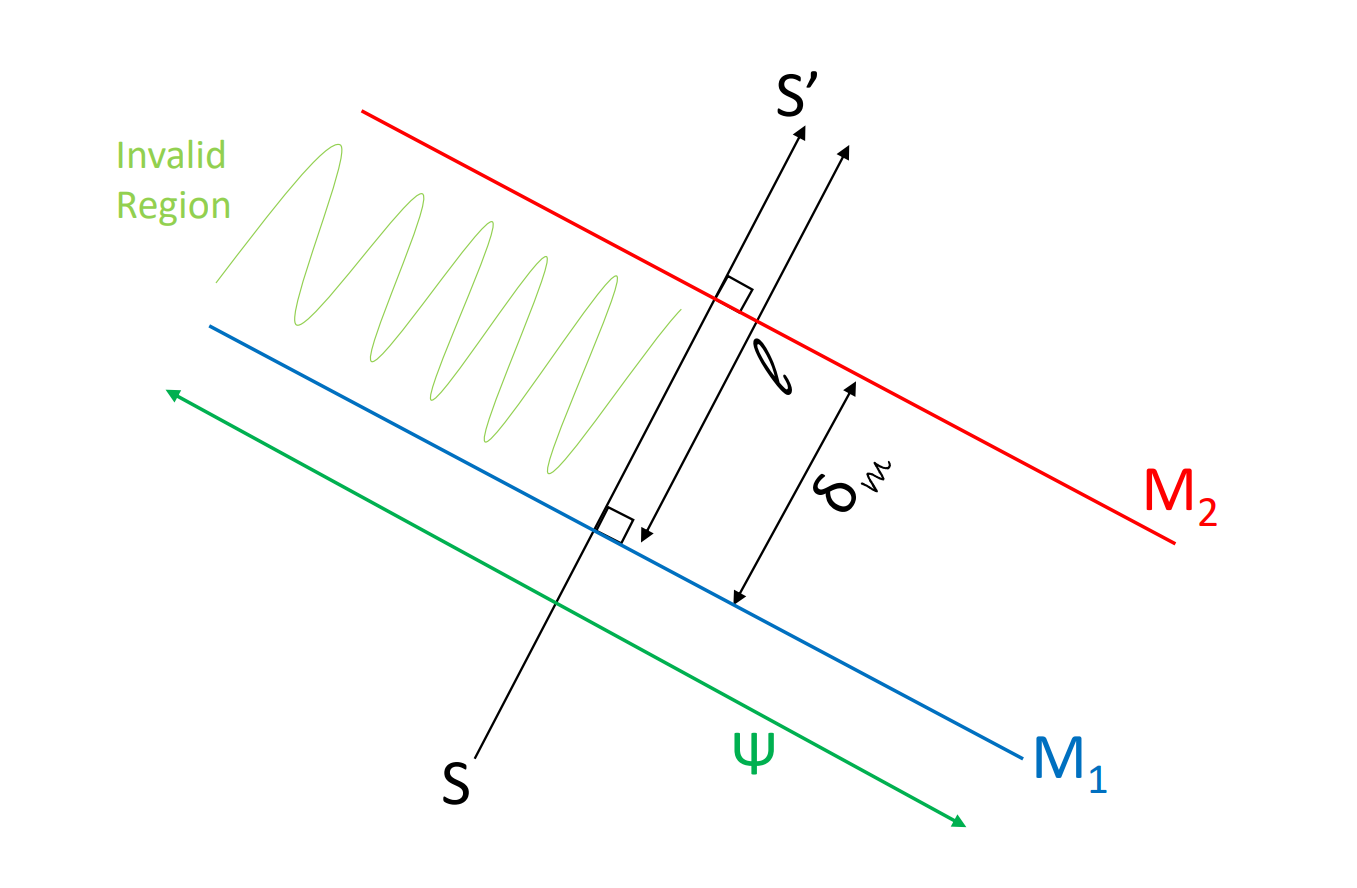}}
\caption{Parallel model perturbation of arbitrary magnitude $\delta_m$ between linear models $\mathcal{M}_1$ and $\mathcal{M}_2$. The range of the data is represented by manifold $\Psi$.}
\label{app:fig:thm2}
\end{center}
\end{figure}

\begin{lemma}
If the model $\mathcal{M}_1$ is linear, then the distribution of the distances between the counterfactuals/recourses $\mathcal{CF}_1$ and the hyperplane of the classifier $\mathcal{M}_1$ follows the exponential distribution for continuous data and geometric distribution for discrete data i.e., 
$f(l) = \rho \cdot e^{-\rho l}$ or $f(l) = (1-\rho)^{l-1} \cdot \rho$, where $l$ is the distance from $\mathcal{S'}$ to $\mathcal{M}_1$ (figure \ref{app:fig:thm2}).
\label{app:lm:distbn}
\end{lemma}
\begin{proof}
By Proposition \ref{app:mdp} above, we know that the recourse search process follows the Markovian Property, and is therefore memoryless. Since this process is memoryless, it follows either exponential (continuous data) or geometric (discrete data) distributions since these are the only two memoryless distributions\footnote{https://en.wikipedia.org/wiki/Memorylessness}. 

Further, since the classifier is linear, we know that an \emph{unrestricted} recourse search, such as that of sparse counterfactual based recourses, will proceed exactly along the normal to the classifier hyperplane (\cite{Ustun_2019}), in the direction of increasingly positive $\mathcal{M}_1$ classification probabilities. 

Using the aforementioned insights and denoting the probability of the counterfactual explanation search ending at any given iteration $t$ with $\mathcal{S'} = \mathcal{S"}_t$ is always constant $\rho$, we can write the distribution of the distances between the counterfactuals/recourses and the hyperplane of the classifier as: 
\begin{align}
    f(l) \sim \rho \cdot e^{-\rho l} \text{ if the data is continuous }\\
    f(l) = (1-\rho)^{l-1} \cdot \rho \text{ if the data is discrete }
\end{align}
\end{proof}

\begin{theorem}
For a given linear model $\mathcal{M}_1$ with recourses $\mathcal{CF}_1$, and a parallel linear model $\mathcal{M}_2$ with arbitrary (constant) perturbation $\delta_m$, the invalidation probabilities $Q{_\mathcal{S'}}$ are $1 - e^{-\rho \delta_m}$ for continuous (numeric) data, and $1 - (1-\rho)^{\delta_m}$ for categorical (ordinal) data.
\label{app:thrm3}
\end{theorem}
\begin{proof}
Let the recourses be distributed according to some unknown arbitrary distribution with density $f(l)$, where $l \in (0, \infty)$ is the normal distance between $\mathcal{M}_1$ and $\mathcal{S'}$. Then, as is clear from the invalid region between the models $\mathcal{M}_1$ and $\mathcal{M}_2$ illustrated in figure \ref{app:fig:thm2}, the probability of invalidation of the recourses $\mathcal{S'}$ is $Q_{\mathcal{S'}} = \frac{1}{\Psi} \int_{\Psi} \left[ \int_0^{\delta_m} f(l) \,dl \right] \, d\psi $. Here $\psi$ is an arbitrary element of the decision boundary, within the data manifold $\Psi$.

We can now combine this result with known distributions from Lemma \ref{app:lm:distbn} to get:
\begin{align}
    Q_{\mathcal{S'}} &= \frac{1}{\Psi} \int_{\Psi} \left[ \int_0^{\delta_m} f(l) \, dl  \right] \, d\psi \\
    &= \frac{1}{\Psi} \int_{\Psi} \left[ \int_0^{\delta_m} \rho e^{-\rho l} \, dl  \right] \, d\psi \\
    &= \frac{1}{\Psi} \int_{\Psi} \left[ 1 - e^{-\rho \delta_m}  \right] \, d\psi 
    \end{align}
    \begin{align}
    &= \frac{1}{\Psi} \int_{\Psi} \, d\psi \cdot \left[ 1 - e^{-\rho \delta_m}  \right] \\
    &= 1 - e^{-\rho \delta_m}
\end{align}

Similarly, for ordinal data this would be $Q_{\mathcal{S'}} = \frac{1}{\Psi} \int_{\Psi} \left[\sum_{l=1}^{\delta_m} (1-\rho)^{l-1}\cdot\rho \right] \, d\psi = 1 - (1-\rho)^{\delta_m}$. Thus, we have characterised the exact invalidation probabilities for parallel linear models $\mathcal{M}_1$ and $\mathcal{M}_2$, with known distance $\delta_m$.
\end{proof}

We now tend to the case of classifiers with non-linear decision boundaries.

\begin{theorem}
For a given nonlinear model $\mathcal{M}_1$ with recourses $\mathcal{CF}_1$, and a parallel nonlinear model $\mathcal{M}_2$ with known constant perturbation $\delta_m$, the lower bound on the invalidation probabilities is achieved exactly when both models $\mathcal{M}_1$ and $\mathcal{M}_2$ are linear.
\end{theorem}
\begin{proof}
We consider a piecewise linear approximation of the nonlinear models, with an arbitrary degree of precision. At each point in the classifier decsion boundary, we consider the piecewise linear approximation to make an angle $\theta$ with a hypothetical hyperplane in the data manifold $\Psi$. We then proceed identically as in the proof for theorem \ref{app:thrm3} above with:
\[ Q_{\mathcal{S'}} = \frac{1}{\Psi \cos \theta} \int_{\Psi} \left[ \int_0^{\delta_m} f(l) \,dl \right] \, d\psi \] 
where the $\cos \theta$ term reflects that the models are locally inclined at an angle $\theta$ from the hyperplane $\psi$.
\[ \frac{dQ_{\mathcal{S'}}}{d\theta} \propto \tan \theta \]
It is easy to see that $Q_{\mathcal{S'}}$ will be maximized when $\theta = 0, \forall \theta$, because $\frac{dQ_{\mathcal{S'}}}{d\theta} \propto \tan \theta = 0 \implies \theta = 0$. If each element of the piecewise linear approximation makes a constant angle of $0$ with the models, then the models themselves must be linear. Thus, the lower bound on invalidation probability for non-linear models must exactly be the invalidation probability for linear models, given the same data (manifold).
\end{proof}


\section{Experimental Analysis}

\subsection{Compute Details and Licenses}

The code and data associated can be found online at: <link hidden for peer-review, code is attached in supplement>. All experiments were run on a single computer with an i-7 8th gen processor and 16 GB of RAM (no GPU was used). The final results table can be generated in $\sim 4$ hours using this setup.

The bail dataset is proprietary, which we obtained from \cite{lakkaraju2016interpretable}. The other datasets used are all in the public domain (creative commons license) \cite{ngcredit, UCI, geospatial}.

\newpage

\subsection{Datasets Overview}

\begin{table}[ht]\centering
\scriptsize
\begin{tabular}{ccc||rr|rr}\toprule
 \multicolumn{3}{c}{\textbf{Dataset}} & \multicolumn{2}{c}{\textbf{\# D1 points}} & \multicolumn{2}{c}{\textbf{\# D2 points}} \\
\textbf{Name} &\textbf{Distribution Shift} &\textbf{Domain} &\textbf{total} &\textbf{negative class} &\textbf{total} &\textbf{negative class} \\\midrule
Bail & Temporal & Criminal Justice & 8395 & 5203 & 8595 & 5430 \\
Schools & Geospatial & Education & 129 & 46 & 122 & 27 \\
German-credit & Data-correction & Finance / Lending & 900 & 275 & 900 & 271 \\
\bottomrule
\end{tabular}
\vspace{0.1in}
\caption{\textbf{Dataset Summary Stats}: Overview of each dataset, including total number of points, and how many of them belonged to the negative class. A perfect classifier would identify all of these (negatively labelled) points as negative, and we'd like to generate recourses for these points.}
\label{app:tab:dsum}
\end{table}

\subsection{Results}


\begin{table}[ht]\centering
\scriptsize
\begin{tabular}{lc||rr|rr|rr}\toprule
\textbf{Algorithm} &\textbf{Model} &\textbf{M1 perf. on D1} &\textbf{M1 perf. on D2} &\textbf{M2 perf. on D2} &\textbf{M2 perf. on D1} &\textbf{CF1 Size} &\textbf{Invalidation \%} \\\midrule \midrule
\multirow{6}{*}{\textbf{AR}} &LR &94 &94 &95.4 &98 &5592 &96.6 \\
&RF &99 &99 &99.5 &98 &4435 &0.05 \\
&XGB &100 &99 &99.7 &96 &1459 &0 \\
&SVM &81 &87 &78.9 &88 &6108 &3.05 \\
&DNN (s) &99 &99 &99.4 &98 &1521 &19.26 \\
&DNN (l) &99 &99 &99.6 &99 &1817 &0 \\
\midrule
\multirow{6}{*}{\textbf{CFE}} &LR &94 &94 &95.4 &98 &5601 &98.29 \\
&RF &99 &99 &99.5 &98 &5196 &0.71 \\
&XGB &100 &99 &99.7 &96 &5187 &0.46 \\
&SVM &81 &87 &78.9 &88 &6108 &100 \\
&DNN (s) &99 &99 &99.4 &98 &4955 &91.38 \\
&DNN (l) &99 &99 &99.6 &99 &763 &0.13 \\
\bottomrule
\end{tabular}
\vspace{0.1in}
\caption{\textbf{Bail} (Temporal Distribution Shifts): Invalidation by model \textbf{M2} of recourses generated using model \textbf{M1}, along with various model accuracies (performance of \textbf{M1} on \textbf{D1} and \textbf{D2}; and of \textbf{M2} on \textbf{D2} and \textbf{D1}), and total recourses generated (\textbf{CF1 Size}). We see that both M1 and M2 have similarly high accuracies on D1 and D2, and yet we observe significant amounts of recourses are being invalidated: for both AR and CFE.}
\label{app:tab:temporal}
\end{table}

\begin{table}[ht]\centering
\scriptsize
\begin{tabular}{lc|rr|rr|rr}\toprule
\textbf{Algorithm} &\textbf{Model} &\textbf{M1 perf. on D1} &\textbf{M1 perf. on D2} &\textbf{M2 perf. on D2} &\textbf{M2 perf. on D1} &\textbf{CF1 Size} &\textbf{Invalidation \%} \\\midrule \midrule
\multirow{6}{*}{\textbf{AR}} &LR &88 &90 &93 &82 &47 &76.6 \\
&RF &89 &91 &92 &89 &0 &\textit{NAN} \\
&XGB &85 &89 &93 &81 &0 &\textit{NAN} \\
&SVM &80 &83 &91 &78 &40 &90 \\
&DNN (s) &83 &86 &87 &70 &0 &\textit{NAN} \\
&DNN (l) &82 &86 &93 &75 &0 &\textit{NAN} \\
\midrule
\multirow{6}{*}{\textbf{CFE}} &LR &88 &90 &93 &82 &47 &65.96 \\
&RF &89 &91 &92 &89 &17 &76.47 \\
&XGB &85 &89 &93 &81 &14 &57.14 \\
&SVM &80 &83 &91 &78 &54 &100 \\
&DNN (s) &83 &86 &87 &70 &2 &50 \\
&DNN (l) &82 &86 &93 &75 &33 &30.3 \\
\bottomrule
\end{tabular}
\vspace{0.1in}
\caption{\textbf{Schools} (Geospatial Distribution Shifts): Invalidation by model \textbf{M2} of recourses generated using model \textbf{M1}, along with various model accuracies (performance of \textbf{M1} on \textbf{D1} and \textbf{D2}; and of \textbf{M2} on \textbf{D2} and \textbf{D1}), and total recourses generated (\textbf{CF1 Size}). We see that the performance of M1 remains high even on D2, but this is not the case for M2 performance on D1, and therefore recourses are sometimes invalidated: for both AR and CFE.}
\label{app:tab:geospatial}
\end{table}

\begin{table}[ht]\centering
\scriptsize
\begin{tabular}{lc|rr|rr|rr}\toprule
\textbf{Algorithm} &\textbf{Model} &\textbf{M1 perf. on D1} &\textbf{M1 perf. on D2} &\textbf{M2 perf. on D2} &\textbf{M2 perf. on D1} &\textbf{CF1 Size} &\textbf{Invalidation \%} \\\midrule \midrule
\multirow{6}{*}{\textbf{AR}} &LR &71 &51 &75 &70 &154 &7.79 \\
&RF &73 &54 &73 &53 &20 &35 \\
&XGB &74 &64 &75 &62 &25 &8 \\
&SVM &63 &70 &69 &56 &1 &100 \\
&DNN (s) &68 &70 &69 &70 &0 &\textit{NAN} \\
&DNN (l) &66 &71 &67 &69 &1 &0 \\
\midrule
\multirow{6}{*}{\textbf{CFE}} &LR &71 &51 &75 &70 &154 &3.9 \\
&RF &73 &54 &73 &53 &258 &36.82 \\
&XGB &74 &64 &75 &62 &253 &23.72 \\
&SVM &63 &70 &69 &56 &1 &0 \\
&DNN (s) &68 &70 &69 &70 &0 &\textit{NAN} \\
&DNN (l) &66 &71 &67 &69 &57 &0 \\
\midrule
\multirow{6}{*}{\textbf{Causal}} &LR &69 &71 &71 &70 &61 &0 \\
&RF &65 &92 &64 &90 &256 &96.09 \\
&XGB &64 &93 &68 &93 &8 &12.5 \\
&DNN (s) &69 &70 &70 &69 &0 &\textit{NAN} \\
&DNN (l) &69 &70 &70 &69 &0 &\textit{NAN} \\
\bottomrule
\end{tabular}
\vspace{0.1in}
\caption{\textbf{German-credit} (Data-correction Distribution Shifts): Invalidation by model \textbf{M2} of recourses generated using model \textbf{M1}, along with various model accuracies (performance of \textbf{M1} on \textbf{D1} and \textbf{D2}; and of \textbf{M2} on \textbf{D2} and \textbf{D1}), and total recourses generated (\textbf{CF1 Size}). We see that model performance after distribution shift is usually slightly worse than on initial distributions, and therefore recourses are sometimes invalidated: for AR, CFE, and Causal recourses.}
\label{app:tab:correction}
\end{table}